\pdfoutput=1
\documentclass[preprint,aoas]{imsart}

\usepackage[T1]{fontenc}

\usepackage{tgtermes}
\usepackage{amssymb}
\usepackage{amsthm,amsmath,natbib}

\usepackage{scalefnt,letltxmacro}
\LetLtxMacro{\oldtextsc}{\textsc}
\renewcommand{\textsc}[1]{\oldtextsc{\scalefont{1.10}#1}}
\usepackage[scaled=0.92]{PTSans}

\usepackage[usenames,dvipsnames]{xcolor}
\definecolor{shadecolor}{gray}{0.9}
\definecolor{myGray}{gray}{0.3}

\usepackage[final,expansion=alltext]{microtype}
\usepackage[english]{babel}
\usepackage[parfill]{parskip}
\usepackage{afterpage}
\usepackage{framed}
\usepackage{nicefrac}

\usepackage{lineno}

\usepackage{ragged2e}

\usepackage{graphicx}
\usepackage{subfig}

\usepackage{booktabs}
\usepackage{multirow}
\usepackage{longtable}
\usepackage{arydshln}

\usepackage[algoruled]{algorithm2e}
\usepackage{listings}
\usepackage{fancyvrb}
\fvset{fontsize=\normalsize}

\usepackage[bookmarks=false,colorlinks,linkcolor=black,citecolor=MidnightBlue,urlcolor=MidnightBlue]{hyperref}

\usepackage[nameinlink]{cleveref}
\creflabelformat{equation}{#1#2#3}
\crefname{equation}{eq.}{eqs.}
\Crefname{equation}{Eq.}{Eqs.}

\usepackage[acronym,smallcaps,nowarn]{glossaries}

\usepackage[colorinlistoftodos,
           prependcaption,
           textsize=small,
           backgroundcolor=yellow,
           linecolor=lightgray,
           bordercolor=lightgray]{todonotes}

\usepackage{framed}

\usepackage{enumitem}

\usepackage[left=1.75in,
            right=1.75in,
            bottom=1in,
            top=1in,
            marginparsep=25pt,
            marginparwidth=2in]{geometry}

\DeclareRobustCommand{\mb}[1]{\ensuremath{\boldsymbol{\mathbf{#1}}}}

\DeclareRobustCommand{\E}[2]{\mathbb{E}_{#1}\left[#2\right]}

\newcommand{\g}{\, | \,}
\newcommand{\prm}{\, ; \,}

\newcommand{\Lcal}{\mathcal{L}}
\newcommand{\Ncal}{\mathcal{N}}

\newcommand{\Tcal}{\mathcal{T}}

\newcommand{\bx}{\mathbf{x}}

\newcommand{\mby}{\mathbf{y}}

\newcommand{\beps}{\boldsymbol{\epsilon}}

\newacronym{ADVI}{advi}{automatic differentiation variational inference}

\newacronym{BBVI}{bbvi}{black-box variational inference}

\newacronym{CDF}{cdf}{cumulative density function}
\newacronym{CTM}{ctm}{correlated topic model}

\newacronym[\glslongpluralkey={deep exponential families}]{DEF}{def}{deep exponential family}
\newacronym{DMIS}{dmis}{deterministic multiple importance sampling}

\newacronym{ELBO}{elbo}{evidence lower bound}
\newacronym{EM}{em}{expectation-maximization}

\newacronym{FA}{fa}{factor analysis}

\newacronym{GNTS}{gn-ts}{gamma-normal time series model}
\newacronym{G-REP}{g-rep}{generalized reparameterization}

\newacronym{HPF}{hpf}{hierarchical {P}oisson factorization}

\newacronym{KL}{kl}{{K}ullback-{L}eibler}

\newacronym{LDA}{lda}{latent {D}irichlet allocation}

\newacronym{MF}{mf}{matrix factorization}
\newacronym{MIS}{mis}{multiple importance sampling}

\newacronym{OBBVI}{o-bbvi}{overdispersed black-box variational inference}

\newacronym{PEMB}{p-efe}{{P}oisson exponential family embeddings}

\newacronym{SVI}{svi}{stochastic variational inference}

\newacronym{tSNE}{t-sne}{t-distributed stochastic neighbor embedding}

\newacronym{UPC}{upc}{universal product code}

\newacronym{VI}{vi}{variational inference}

\newcommand{\shopper}{\textsc{shopper}}

\startlocaldefs
\DeclareMathAlphabet{\mathpzc}{OT1}{pzc}{m}{it}
\DeclareFontFamily{OT1}{pzc}{}
\DeclareFontShape{OT1}{pzc}{m}{it}{<-> s * [1.20] pzcmi7t}{}
\DeclareMathAlphabet{\mathpzc}{OT1}{pzc}{m}{it}
\endlocaldefs

\begin{document}

\begin{frontmatter}

\title{SHOPPER: A Probabilistic Model of Consumer Choice with Substitutes and Complements}
\runtitle{SHOPPER: A Probabilistic Model of Consumer Choice}

\begin{aug}
\author{\fnms{Francisco J.\ R.} \snm{Ruiz}\corref{}\thanksref{m1,m2}\ead[label=e1]{f.ruiz@columbia.edu}},
\author{\fnms{Susan} \snm{Athey}\thanksref{m3}\ead[label=e2]{athey@susanathey.com}}
\and
\author{\fnms{David M.} \snm{Blei}\thanksref{m2}\ead[label=e3]{david.blei@columbia.edu}}

\runauthor{F.\ Ruiz et al.}

\affiliation{University of Cambridge,\thanksmark{m1} Columbia University,\thanksmark{m2} and Stanford University\thanksmark{m3}}



\end{aug}

\renewcommand{\abstractname}{}
\begin{abstract}
  We develop \shopper, a sequential probabilistic model of {shopping
  data}.  \shopper\ uses interpretable components to model the
  forces that drive how a customer chooses products; in particular, we
  designed \shopper\ to capture how items interact with other items.
  We develop an efficient posterior inference algorithm to estimate
  these forces from large-scale data, and we analyze a large dataset
  from a major chain grocery store.  We are interested in answering
  counterfactual queries about changes in prices.  We found that
  \shopper\ provides accurate predictions even under price
  interventions, and that it helps identify complementary and
  substitutable pairs of products.
\end{abstract}



\end{frontmatter}




\section{Introduction}
\label{sec:introduction}

\begin{figure}[t]
  \centering
  \includegraphics[width=0.7\textwidth]{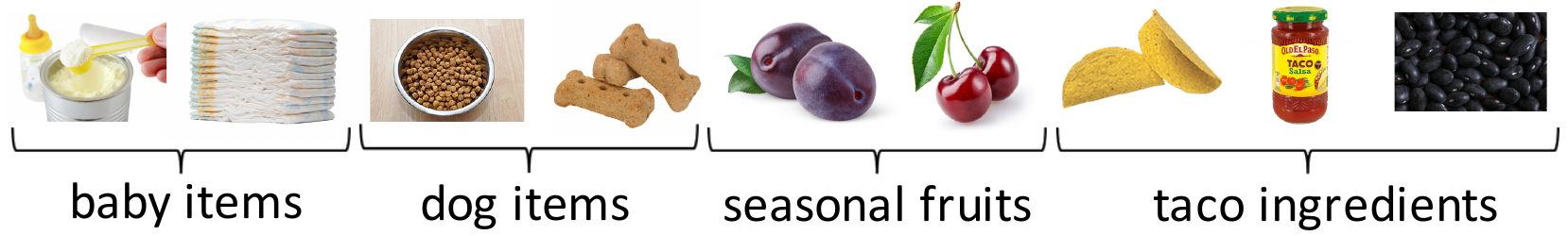}
  \caption{An example of a shopping cart, generated under several
  underlying interrelated forces (customer needs, availability of
  seasonal fruits, complementary sets of items).
  \label{fig:sample_shopping_cart}}
\end{figure}

Large-scale shopping cart data provides unprecedented opportunities
for researchers to understand consumer behavior and to predict how it
responds to interventions such as promotions and price changes.
Consider the shopping cart in \Cref{fig:sample_shopping_cart}. This customer has purchased
items for their baby (diapers, formula), their dog (dog food, dog
biscuits), some seasonal fruit (cherries, plums), and the ingredients
for tacos (taco shells, salsa, and beans). Shopping cart datasets may
contain thousands or millions of customers like this one, each
engaging in dozens or hundreds of shopping trips.

In principle, shopping cart datasets could help reveal important
economic quantities about the marketplace.  They could help evaluate
counterfactual policies, such as how changing a price of a product
will affect the demand for it and for other related products; and they
could help characterize consumer heterogeneity, which would allow
firms to consider different interventions for different segments of
their customers.  However, large-scale shopping cart datasets are too
complex and heterogenous for classical methods of analysis, which
necessarily prune the data to a handful of categories and a small set
of customers.  In this paper, our goal is to develop a method that can
usefully analyze large collections of complete shopping baskets.

Shopping baskets are complex because, as the example shopper in \Cref{fig:sample_shopping_cart}
demonstrates, many interrelated forces are at play when determining
what a customer decides to buy. For example, the customer might
consider how well the items go together, their own personal
preferences and needs (and whims), the purpose of the shopping trip,
the season of the year, and, of course, the prices and the customer's
personal sensitivity to them.  Moreover, these driving forces of
consumer behavior are unobserved elements of the market.  Our goal is
to extract them from observed datasets of customers' final purchases.

To this end we develop \shopper, a sequential probabilistic model of
market baskets.  \shopper\ uses interpretable components to model the
forces that drive customer choice, and we designed \shopper\ to
capture properties of interest regarding how items interact with other
items; in particular, we are interested in answering counterfactual
queries with respect to item prices.  We also develop an efficient
posterior inference algorithm to estimate these forces from
large-scale data.  We demonstrate \shopper\ by analyzing data from a
major chain grocery store \citep{Che2012}.
We found that \shopper\ provides accurate
predictions even under price interventions, and that it also helps
identifying complementary and substitutable pairs of items.



\subsection{Main idea}

\shopper\ is a hierarchical latent variable model of market baskets
for which the generative process comes from an assumed model of
consumer behavior. In the language of social science,
  \shopper\ is a structural model of consumer behavior, where the
  elements of the model include the specification of consumer
  preferences, information, and behavior (i.e., utility maximization).
  In the language of probabilistic models, it can equivalently be
  defined by a generative model.

\shopper\ posits that a customer
walks into the store and chooses items sequentially.  Further, the
customer might decide to stop shopping and pay for the items; this is
the ``checkout'' item.  At each step, the customer chooses among the
previously unselected items, while conditioning on the items already
in the basket.  The customer's choices also depend on various other
aspects: the prices of the items and the customer's sensitivities to
them, the season of the year, and the customer's general shopping
preferences {(which are specific for each customer)}.

One key feature of \shopper\ is that each possible item is associated
with latent attributes {$\alpha_c$}, vector representations that are learned from
the data.  This is similar in spirit to methods from machine learning
that estimate semantic attributes of vocabulary words by analyzing the
words close to them in sentences~\citep{Bengio2003}.  In \shopper\,
``words'' are items and ``sentences'' are baskets of purchased items.

\shopper\ uses the latent attributes in two ways.  First, they help
represent the existing basket when considering which item to select
next. Specifically, each possible item is also associated with a
vector of interaction coefficients {$\rho_c$}, parameters that represent which
kinds of basket-level attributes it tends to appear with.  For example, the
interaction coefficients and attributes can capture that when taco
shells are in the basket, the customer has a high probability of
choosing taco seasoning.  Second, they are used as the basis for
representing customer preferences. Each customer in the population
also has a vector of interaction coefficients {$\theta_u$}, which we call
``preferences,'' that represent the types of items that they tend to
purchase.  For example, the customer preferences and attributes can capture
that some customers tend to purchase baby items or dog food.

{Mathematically, at the $i$th step of the sequential process, the customer
chooses item $c$ with probability that depends on the latent features of
item $c$, on her preferences $\theta_u$, and on the representation of the
items that are already in the basket, $\sum_{j=1}^{i-1} \alpha_{y_j}$
{(where $y_j$ indicates the item purchased at step $j$)}.
In its vanilla form, }\shopper{ posits that this probability takes a
log-bilinear form,} 
\begin{align}
  \textrm{Prob}(\textrm{item } c \g \textrm{items in basket}) \propto
  \exp\left\{ \theta_{u}^\top\alpha_c +\rho_c^\top \left(\frac{1}{i-1}
  \sum_{j=1}^{i-1} \alpha_{y_{j}} \right)  \right\}.
\end{align}
\Cref{sec:model}{ provides more details about the model and also
describes how to incorporate price and seasonal effects.}

\begin{figure}[t]
  \centering
  \includegraphics[width=0.6\textwidth]{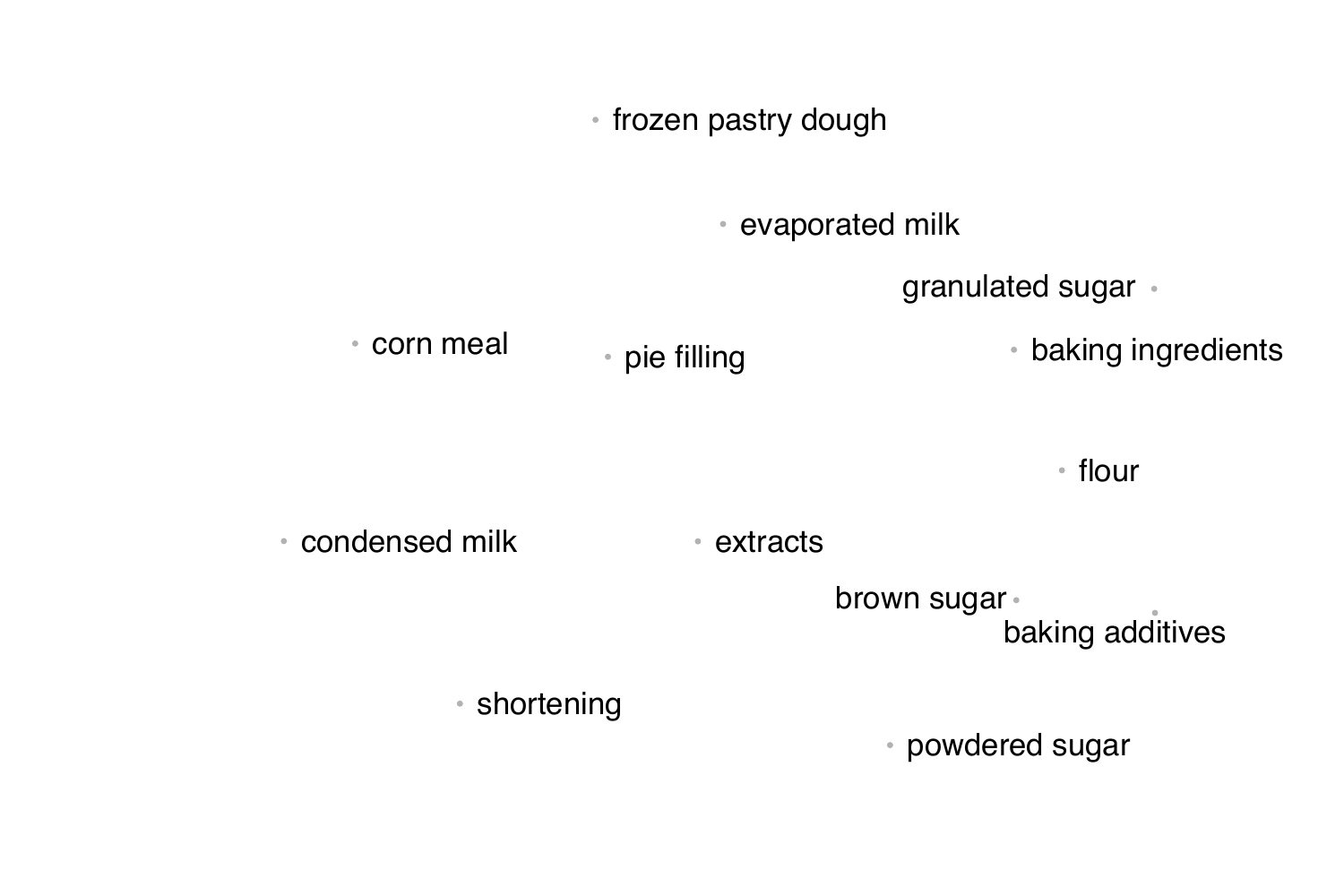}
  \caption{{A region of the two-dimensional projection of the learned item
  attributes, which corresponds to baking ingredients and additives.
  Items that are similar have representations that are close together.}
  \label{fig:tsne_example}}
\end{figure}

When learned from data, the latent attributes {$\alpha_c$} capture meaningful
dimensions of the data. For example, \Cref{fig:tsne_example} illustrates a
two-dimensional projection of learned latent item attributes.  Similar
items are close together in the attribute space, even though explicit
attributes (such as category or purpose) are not provided to the
algorithm.



In the simplest \shopper\ model, the customer is myopic: at each stage
they do not consider that they may later add additional items into the
basket.  This assumption is problematic when items have strong
interaction effects, since it is likely that a customer will consider
other items that complement the currently chosen item.  For example,
if a customer considers purchasing taco seasoning, they should also
consider that they will later want to purchase taco shells.

To relax this assumption, we include
a second key feature of \shopper, called ``thinking
ahead,'' where the customer considers the next choice in making the
current choice.  For example, consider a scenario where taco shells
have an unusually high price and where the customer is currently
contemplating putting taco seasoning in the basket.  As a consequence
of thinking ahead, the model dampens the probability of choosing
seasoning because of the high price of its likely complement, taco
shells.  \shopper\ models one step of thinking ahead.  (It
may be plausible in some settings that customers think further ahead
when shopping; we leave this extension for future work.)

\subsection{Main results}

We fit \shopper\ to a large data set of market baskets, estimating the
latent features of the items, seasonal effects, preferences of the
customers, and price sensitivities. We evaluate the approach with
held-out data, used in two ways.  First, we hold out items from the
data at random and assess their probability with the posterior
predictive distribution.  This checks how well \shopper\ (and our
corresponding inference algorithm) captures the distribution of the
data.  Second, we evaluate on held-out baskets in which items have
larger price variation compared to their average values.  This
evaluation assesses how well \shopper\ can evaluate counterfactual
price changes.  With both modes of evaluation, we found \shopper\ gave
better predictions than state-of-the-art econometric choice models and
machine learning models for item consumption{, which are based on
either user/item factorization, such as factor analysis or Poisson
factorization }\citep{Gopalan2015}{, or on item/item factorization, such
as exponential family embeddings }\citep{Rudolph2016}.

Further, \shopper\ can help identify pairs of items that are
substitutes and complements---where purchasing one item makes the
second item less or more attractive, respectively.  This is one of the
fundamental challenges of analyzing consumer behavior data.
{Formally, two items are complements if the demand of one is increased when
the price of the other decreases; they are substitutes if the demand for one
rises when the price of the other item increases.
Studying complementary and substitutable items is key to make predictions
about the (joint) demand.}

Notice that being complements is more than the propensity for two
items to be co-purchased; items may be purchased together for many
reasons.  As one reason, two items may be co-purchased because people
who like one item tend to also like the other item.  For example, baby
formula and diapers are often co-purchased but they are not
complements---when baby formula is unavailable at the store, it does
not affect the customer's preference for diapers.

\shopper\ can disentangle complementarity from other sources of
co-purchase because prices change frequently in the dataset.  If two
items are co-purchased but not complements, increasing the price of
one does not decrease the purchase rate of the other. Using the
item-to-item interaction term in \shopper, we develop a measure to
quantify how complementary two items are.  (In this sense, \shopper\
corroborates the theory of \citet{athey1998empirical} among others,
who show that if there is sufficient variation in the price of each
item then it is theoretically possible to separate correlated
preferences from complementarity.)

Now we turn to substitutes, where purchasing one item decreases the
utility of another, e.g., two brands of (otherwise similar) taco
shells.  Although in principle substitutes can be treated
symmetrically to complements (increasing the price of one brand of
taco shells increases the probability that the other is purchased if
they are substitutes), in practice the two concepts are not equally
easy to discover in the data.  The difference is that in most shopping
data, purchase probabilities are very low for all items, and so most
pairs of items are rarely purchased together.  In this paper, we
introduce an alternative way to find relationships among products,
``exchangability.''  Two products are exchangeable if they tend to
have similar pairwise interactions with other products; for example,
two brands of taco shells would have similar interactions with
products such as tomatoes and beans.  Of course, products that are
usually purchased together, like hot dogs and buns, will also have
similar pairwise interactions with other products (such as ketchup).
Our results suggest that items that are exchangeable and not
complementary tend to be substitutes (in the sense of being in the
same general product category in the grocery store hierarchy).



\section{A Bayesian model of sequential discrete choice}
\label{sec:model}



We develop \shopper, a sequential probabilistic model of market
baskets.  \shopper\ treats each choice as one over the set of
available items, where the attributes of each item are latent
variables.

We describe \shopper\ in three stages. \Cref{sec:model_sequence}
describes a basic model of sequential choice with latent item
attributes; \Cref{sec:model_eta} extends the model to capture user
heterogeneity, seasonal effects, and price; \Cref{sec:model_lookahead}
develops ``thinking ahead,'' where we model each choice in a way that
considers the next choice as well.

\shopper\ comes with significant computational challenges, both
because of its complex functional form and the size of the data that
we would like to analyze.  We defer these challenges to
\Cref{sec:inference}, where we develop an efficient variational
algorithm for approximate posterior inference.

\subsection{Sequential choice with latent item attributes}

\label{sec:model_sequence}

We describe the basic model in terms of its generative process or, in
the language of economics and marketing, in terms of its structural
model of consumer behavior.  Each customer walks into the store and
picks up a basket.  She chooses items to put in the basket, one by
one, each one conditional on the previous items in the basket.  The
process stops when she purchases the ``checkout'' item, which we treat
as a special item that ends the trip.

From the utility-maximizing perspective, \shopper\ works as follows.
First, the customer walks into the store and obtains utilities for
each item.  She considers all of the items in the store and places the
highest-utility item in her basket. In the
second step, once the first item is in the basket, the customer again
considers all remaining items in the store and selects the highest-utility
choice.  However, relative to the first decision, the utilities of the
products change.  First, the specification of the utility allows for
interaction effects---items may be substitutes (e.g., two brands of
taco seasoning) or complements (e.g., taco shells and taco seasoning), in the
sense that the utility of an item may be higher or lower as a result
of having the first item in the basket.  Second, the customer's
utilities have a random component that changes as she shops.  This
represents, for example, changing ideas about what is needed, or
different impressions of products when reconsidering them.
(Another extension of the model would be to allow for
  correlation of the random component over choice events within a
  shopping trip; we leave this for future work.)  The customer repeats
this process---adjusting utilities and choosing the highest-utility
item among what is left in the store---until the checkout item is the
highest-utility item.

{In some applications, it
may not be reasonable to model the consumer as considering all
items.  In a supermarket, a customer might consider one part of the
store at the time; in an online store, a customer may only consider
the products that appear in a search results page.  Corresponding
extensions to }\shopper{ are straightforward but, for simplicity, we
do not include them here.  Modeling customers as choosing the most
desirable items in the store first ensures that, among goods that
are close substitutes, they select the most desirable one.}





We describe {the sequential choice model} more formally.  Consider the $t$th trip
and let $n_t$ be the number of choices, i.e., the number of purchased items.  Denote
the (ordered) basket $\mby_t = (y_{t1}, \ldots, y_{tn_t})$, where each
$y_{ti}$ is one of $C$ items and choice $y_{tn_t}$ is always the
checkout item. 
Let $\mby_{t,i-1}$ denote the items that
are in the basket up to position $i$,
$\mby_{t,i-1} = (y_{t1}, \ldots, y_{t,i-1})$.

Consider the choice of the $i$th item.  The customer makes this choice
by selecting the item $c$ that has maximal utility
$U_{t,c}(\mby_{t,i-1})$, which is a function of the items in the
basket thus far.  The utility of item $c$ is
\begin{align}
  U_{t,c}(\mby_{t,i-1})=\Psi(c, \mby_{t,i-1})+\epsilon_{t,c}.
\end{align}
Here $\Psi(c, \mby_{t, i-1})$ is the deterministic part of the
utility, a function of the other items in the basket. 
We define $\Psi(c,\mby_{t,i-1}) \equiv -\infty$ for $c \in \mby_{t,i-1}$,
so that the choice is effectively over the set of non-purchased items.
The random variable $\epsilon_{t,c}$ is assumed to follow a zero-mean
Gumbel distribution (generalized extreme value type-I), which is independent
across items.

This behavioral rule---and particularly the choice of Gumbel error
terms---implies that the conditional choice probability of an item $c$
that is not yet in the basket is a softmax,
\begin{align}
  \label{eq:choice-prob}
  p(y_{ti} = c \g \mby_{t,i-1}) =
  \frac{\exp\{\Psi(c, \mby_{t,i-1})\}}
  {\sum_{c' \not \in \mby_{t,i-1}} \exp\{\Psi(c', \mby_{t,i-1})}.
\end{align}
Notice the denominator is a sum over a potentially large number of
items.  For example, \Cref{sec:experiments} analyzes data with nearly
six thousand items. \Cref{sec:inference} describes fast methods for handling
this computational bottleneck.

\Cref{eq:choice-prob} gives a model of observed shopping trips, from
which we can infer the utility function.  The core of \shopper\ is in
the unnormalized log probabilities $\Psi(c, \mby_{t,i-1})$, which
correspond to the mean utilities of the items.  We assume that they
have a log-linear form,
\begin{align}
  \label{eq:log-linear}
  \Psi(c, \mby_{t,i-1}) = \psi_{tc}+\rho_c^\top \left(\frac{1}{i-1}
  \sum_{j=1}^{i-1} \alpha_{y_{tj}} \right).
\end{align}
There are two terms.  The first term $\psi_{tc}$ is a latent variable
that varies by item and by trip; below we use this term to capture properties such as
user heterogeneity, seasonality, and price.  We focus now on the second term,
which introduces two important latent variables: the per-item
\textit{interaction coefficients} $\rho_c$ and the \textit{item
  attributes} $\alpha_c$, both real-valued $K$-vectors.  When
$\rho_c^\top \alpha_{c'}$ is large, then having $c'$ in the basket
increases the benefit to the customer of placing $c$ in the basket
(the items are complements in the customer's utility); conversely,
when the expression is negative, the items are substitutes.

Unlike traditional factorization methods, the factorization is
asymmetric.  We interpret $\alpha_c$ as latent item characteristics or
attributes; we interpret $\rho_c$ as the interaction of item $c$ with
other items in the basket, as described by their attributes (i.e.,
their $\alpha$'s).  These interactions allow that even though two
items might be different in their latent attributes (e.g., taco shells and
beans), they still may be co-purchased because they are
complements to the consumer.  It can also capture that similar items
(e.g., two different brands of the same type of taco seasoning) may
explicitly \textit{not} be purchased together---these are
substitutable items.  Below we also allow that the latent attributes $\alpha_c$
further affect the item's latent mean utility on the trip
$\psi_{tc}$.

We also note that the second term has a scaling factor, $1/(i-1)$.  This
scaling factor captures the idea that in a large basket, each individual item
has a proportionally smaller interaction with new purchases.  This may
be a more reasonable assumption in some scenarios than others, exploring
alternative ways to account for basket size is a direction for future work.  



Finally, note we assumed that $\Psi(\cdot)$ is additive in the other items in
the basket. {With an additive model, the interaction term
$\rho_c^\top \alpha_{c^\prime}$ affects the probability of item $c$ in the same
way for each item $c^\prime$ in the basket. In addition, this choice }rules out
non-linear interactions with other items. Again, this restriction may be more
realistic in some applications than in
others, but it is possible to extend the model to consider more complex
interaction patterns if necessary. In this paper, we only consider linear
interaction effects.


\subsubsection{Baskets as unordered set of items}  Given
\Cref{eq:choice-prob}, the probability of an ordered basket is the
product of the individual choice probabilities.  Assuming that
$y_{tn_t}$ is the checkout item, it is
\begin{align}
  \label{eq:p_y_noperm}
  p(\mby_t \g \rho, \alpha) = \prod_{i=1}^{n_t}
  p(y_{ti} \g \mby_{t,i-1}, \rho, \alpha).
\end{align}
The probabilities come from \Cref{eq:choice-prob} and
\Cref{eq:log-linear}, and we have made explicit the dependence on the
interaction coefficients and item attributes.  The parameters of this
softmax are determined by the interaction vectors $\rho_c$ and the
attributes $\alpha_c$ of the items that have already been
purchased. Given a dataset of (ordered) market baskets, we can use
this likelihood to fit each item's latent attributes and interaction.

In many datasets, however, the order in which the items are added to
the basket is not observed.  \shopper\ implies the likelihood of an
unordered set $\mathpzc{y}_t$ by summing over all possible orderings,
\begin{align}
 \label{eq:p_y_perm}
  p(\mathpzc{y}_t \g \rho,\alpha) = \sum_{\pi} p(\mby_{t, \pi}\g \rho,\alpha).
\end{align}
Here $\pi$ is a permutation (with the checkout item fixed to the last
position) and $\mby_{t,\pi}$ is the permuted basket
$(y_{t,\pi_1}, \ldots, y_{t,\pi_{n_t}})$. Its probability is in
\Cref{eq:p_y_noperm}.  In \Cref{sec:experiments}, we study a large
dataset of unordered baskets; this is the likelihood that we use when
fitting \shopper.



\subsubsection{Utility maximization of baskets}

\label{sec:basket_utility}

Here we describe how the sequential model behind \shopper\ relates to
a utility maximization model over unordered sets of items, in which
the customer has a specific utility for each basket (i.e., with $2^C$
choices, where $C$ is the number of items).

Let $\mathpzc{y}_t$ be the (unordered) set of items purchased in trip
$t$.  Define a consumer's utility over unordered baskets as follows:
\begin{align}
\label{eq:unorderedU}
  \widetilde{U}_t(\mathpzc{y}_t) = \sum_{c \in
  \mathpzc{y}_t}{\psi_{tc}}+ \frac{1}{|\mathpzc{y}_t|-1}\sum_{(c,c')\in \mathpzc{y}_t \times
  \mathpzc{y}_t:c' \ne c}{\nu_{c,c'}},
\end{align}
where $\nu_{c,c'}$ is a term that describes the interaction between
$c$ and $c'$. Now consider the following model of utility maximization.
At each stage where a consumer must select an item, the consumer has an
extreme form of myopia, whereby she assumes that she will immediately
check out after selecting this item, and she does not consider the
possibility that she could put any items in her basket back on the
shelf.  Other than this myopia, she behaves rationally, maximizing her
utility over unordered items as given by \Cref{eq:unorderedU}.

The behavioral model of sequential shopping is consistent with the
myopic model of utility maximization if (and only if)
$\nu_{c,c^\prime}=\rho_c^\top \alpha_{c'} =\rho_{c'}^\top \alpha_c$;
this says that the impact of product $c$ on the purchase of product
$c'$ is symmetric to the impact of $c$ on product $c'$, holding fixed
the rest of the basket. (We do not impose such a symmetry constraint
in our model because of the reasons outlined in \Cref{eq:log-linear}.)
Thus, we can think of utility maximization by this type of myopic
consumer as imposing an additional constraint on the probabilistic
model.  This follows the common practice in economic modeling to
estimate the richer models motivated by theory, but without imposing
all the restrictions on the parameters implied by that theory; this
approach often simplifies computation.  \citet{browning1991effects}
also estimate an econometric model without imposing symmetry
restrictions implied by utility maximization, and then impose the
restrictions in a second step using a minimum-distance approach.

Finally, we note that a fully rational consumer with full information of the price of all products could in principle consider all of the possible bundles in the store simultaneously and maximize over them. However, given that the number of bundles is $2^C$, we argue that considering such a large number of bundles simultaneously is probably not a good approximation to human behavior. Even if consumers are not as myopic as we assume, it is more realistic to assume that they follow some simple heuristics.

\subsection{Preferences, seasons, popularity, and price}
\label{sec:model_eta}

The basic \shopper\ model of the previous section captures a
customer's sequential choices as a function of latent attributes and
interaction coefficients.  \shopper\ is flexible, however, in that we
can include other forces in the model of customer choice;
specifically, they can be incorporated into the (unobserved) mean utility of each
item $c$, $\psi_{tc}$, which varies with trip and customer. Here we
describe extensions to capture item popularity, customer preferences,
price sensitivity, and seasonal purchasing habits (e.g., for holidays
and growing seasons).  All these factors are important when modeling
real-world consumer demand.

\subsubsection{Item popularity} We capture overall (time-invariant)
item popularity with a latent intercept term $\lambda_c$ for each
item.  When inferred from data,
popular items will have a high value of $\lambda_c$, which will
generally increase their choice probabilities.

\subsubsection{Customer preferences} In our data, each trip $t$ is
associated with a particular customer $u_t$, and that customer's
preferences affect her choices.  We model preference with a
per-customer latent vector $\theta_u$. For each
choice, we add the inner product $\theta_u^\top \alpha_c$ to the
unnormalized log probability of each item.  This term increases the
probability of types of items that the customer tends to purchase.
{Recall that $\theta_u$ is a per-customer variable that appears
in every shopping trip $t$ involving customer $u$.}
Note that \shopper\ shares the attributes $\alpha_c$ with the part of
the model that characterizes interaction effects. The inference algorithm
finds latent attributes $\alpha_c$ that interact with both customer
preferences and also with other items in the basket.


\subsubsection{Price sensitivity} We next include per-customer price
sensitivity.  Let $r_{tc}$ denote the price of item $c$ at trip
$t$. We consider each customer has an individualized price sensitivity
to each item, denoted $\tau_{uc}$, and we add the term
$-\tau_{uc} \log r_{tc}$ to the unnormalized log probabilities in
\Cref{eq:log-linear}.  We place a minus sign in the price term to make
the choice less likely as the price $r_{tc}$ increases.  Further, we
constrain $\tau_{uc}$ to be positive; this constraint ensures that the
resulting price elasticities are negative. (The price
  elasticity of demand $\varepsilon$ is a measure used in economics
  to show the responsiveness of the quantity demanded of a good $y$
  to a change in its price $r$; it is defined as
  $\varepsilon = \partial \log y / \partial \log r$.)

Including $\tau_{uc}$ posits a large number of latent variables (for
many customers and items) and, moreover, it is reasonable to assume
the sensitivities will be correlated, e.g., a customer's sensitivity
to peanut butter and almond butter might be similar.  Thus we use a
matrix factorization to model the price sensitivities.  Specifically,
we decompose the user/item price sensitivity matrix into per-user
latent vectors $\gamma_u$ and per-item latent vectors $\beta_c$, where
$\tau_{uc} = \gamma_{u}^\top \beta_c$.  This factorization models the
complete matrix of price sensitivities with fewer latent variables.

Finally, instead of the raw price $r_{tc}$, we use the normalized price, i.e.,
the price for this trip divided by the per-item mean price.
Normalized prices have two attractive properties. First, they allow
$\beta_c$ to be on a comparable scale, avoiding potential issues that
arise when different items have prices that vary by orders of
magnitude.  Second, they ensure that
the other parameters in \Cref{eq:log-linear} capture the average outcome distribution: the
price term vanishes when the price takes its average value because the
log of the normalized price is zero.

\subsubsection{Seasonal effects} We complete the model with seasonal
effects. \shopper\ is designed to estimate the effect of
counterfactual changes in policy, such as prices.  So, it is important
that the parameters of the model associated with price represent the
true causal effect of price on customer choices.  Seasonal effects are
a potential confounder---the season simultaneously affects the price
and the demand of an item---and so neglecting to control for
seasonality can lead to estimates of the latent variables that cannot
isolate the effects of changing prices.  For example, the demand for
candy goes up around Halloween and the supermarket may decide to put
candy on sale during that period. The demand increases partly because
of the price, but also because of the time of year.  Controlling for
seasonal effects isolates the causal effect of price.

We assume that seasonal effects are constant for each calendar
week and, as for the price effect, we factorize the week-by-item seasonal
effects matrix. 
(This is motivated by \citet{athey2017counterfactual}, who conduct a
series of empirical studies that support the idea that controlling
for week effects is sufficient to identify the causal effect of
price.)
Denote the week of trip $t$ as $w_t \in \{1, \ldots, 52\}$.
We posit per-week latent vectors $\delta_w$ and per-item latent
vectors $\mu_c$, and we include a term in
\Cref{eq:log-linear} that models the effect of week $w$ on item $c$'s
probability, $\delta_w^\top \mu_c$.  Note this allows correlated
seasonal effects across items; items with similar $\mu_c$ vectors will
have similar seasonal effects.  So Halloween candy and pumpkin cookies
might share similar $\mu_c$ vectors.

\subsubsection{Putting it together} We described popularity,
preferences, price, and season.  We combine these effects in the
per-item per-trip latent variable $\psi_{tc}$,
\begin{align}
  \label{eq:psi_tc}
  \psi_{tc} = \underbrace{\lambda_c}_{\textrm{item popularity}} +
  \underbrace{\theta_{u_t}^\top \alpha_c}_{\textrm{customer preferences}} - \underbrace{\gamma_{u_t}^\top\beta_c \log r_{tc}}_{\textrm{price effects}} + \underbrace{\delta_{w_t}^\top \mu_c,}_{\textrm{seasonal effects}}
\end{align}
{where the subscripts $u_t$ and $w_t$ indicate, respectively,
the customer and week corresponding to shopping trip $t$.}
The mean utility $\psi_{tc}$ is used in the probabilities for all $n_t$ choices of
trip $t$.  With these extensions we have introduced several new latent
variables to infer from data: item popularities $\lambda_c$, customer
preferences $\theta_u$, price sensitivity factorizations $\gamma_u$
and $\beta_c$, and seasonal factorizations $\delta_w$ and $\mu_c$.




\subsection{Thinking ahead}
\label{sec:model_lookahead}

As a final extension, we develop a model of customers that ``think
ahead.''  Consider a pair of items $c$ and $c'$, where both
$\rho_c^\top \alpha_{c'}$ and $\rho_{c'}^\top \alpha_{c}$ are high, so
that the value of $c$ increases when $c'$ is in the basket and vice
versa.  In this case, we say that the goods are complements to the
consumer.  Accurately accounting for complementarity is particularly
important when estimating counterfactuals based on different prices;
theoretically increasing the price of one item should lower the
probability of the other.

In this scenario, our baseline model specifies that the consumer will
consider the goods one at a time. The effect of a high price for
item $c'$ is that it reduces the probability the item is chosen (and
if chosen, it is less likely to be chosen early in the ordering).  Not
having item $c'$ in the basket at each new consideration event reduces
the attractiveness of item $c$.  However, a more realistic model has
the consumer anticipate when she considers item $c$ that she might
also consider item $c'$. When the consumer considers the pair, the
high price of one item deters the purchase of both.

We address this issue by introducing \textit{thinking ahead}.  When
considering the item at step $i$, the consumer looks ahead at the
$i+1$th step the customer may purchase.  We emphasize that the model
does not, at this point, assume that the consumer actually purchases
the next item according to that rule.  When the consumer comes to
consider what to choose for step $i+1$, she follows the same
think-ahead-one-step rule just described.

Thinking ahead adjusts the unnormalized log probabilities of the
current item in \Cref{eq:log-linear}.  Specifically, it adds a term to
the log linear model that is the utility of the optimal next item.
(Note this could be the checkout item.)  The unnormalized log
probability is
\begin{align}
  \label{eq:psi_thinkahead}
  \begin{aligned}
  \Psi(c, \mby_{t,i-1}) = & \psi_{tc} + \rho_c^\top \left(\frac{1}{i-1}
  \sum_{j=1}^{i-1} \alpha_{y_{tj}} \right)\\
&\qquad + \max_{c' \not \in
  [\mby_{t,i-1}, c]} \left\{\psi_{tc'} + \rho_{c'}^\top
  \left(\frac{1}{i} \left(\alpha_{c} + \sum_{j=1}^{i-1} \alpha_{y_{tj}} \right)\right)\right\},
\end{aligned}
\end{align}
where $[\mby_{t,i-1}, c]$ is a hypothetical basket that contains the
first $i-1$ purchased items and item $c$. Since the customer has in
mind a tentative next item when deciding about the current item, we
call this ``thinking one step ahead.''

Note that thinking ahead assumes that the item $c^\prime$ will itself be
selected without looking further ahead, that is, it is chosen
purely based on its own utility without accounting for the interaction
effects of $c^\prime$ with future purchases.  The thinking-ahead idea can
be extended to two or more steps ahead, but at the expense of a higher computational
complexity.
See \Cref{sec:illustrative} for an illustrative example that clarifies
how the thinking-ahead procedure works.

The thinking-ahead mechanism is an example of inferring agent's preferences
from their behavior in a dynamic model, a type of exercise with a long history
in applied econometrics (e.g. \citet{wolpin1984estimable, hotz1993conditional}).  From the machine learning perspective, it closely resembles the motivation behind
inverse reinforcement learning \citep{Russell1998,Ng2000}. Inverse
reinforcement learning analyzes an agent's behavior in a variety of
circumstances and aims to learn its reward function.  (This is in
contrast to reinforcement learning, where the reward function is
given.)  To see the connection, consider the unnormalized log
probability of \Cref{eq:psi_thinkahead} as the reward for choosing $c$ and
then acting optimally one step (i.e., choosing the best next
item). With this perspective, fitting \shopper\ to observed data is
akin to learning the reward function.

Finally, note the thinking-ahead model also connects to utility maximization
over unordered baskets, with the utility function given in
\Cref{eq:unorderedU}.  Now the model is consistent with a consumer
who maximizes her utility over unordered items at each step, but where
at each step she is myopic in two ways.  First, as before, she does
not consider that she can remove items from her cart.  Second, she
assumes that she will buy exactly one more item in the store and then
check out, and the next item will be the one that maximizes her utilty
over unordered baskets (taking as given her cart at that point,
including the latest item added, as well as the belief that she will
then check out).  Again, our probabilistic model is consistent
with this interpretation if and only if
$\rho_{c'}^\top \alpha_c = \rho_{c}^\top \alpha_{c'}$ for all
$(c,c')$ {(a constraint that we do not enforce)}.

\subsection{Full model specification}
\label{sec:model_full}

We specified \shopper, a sequential model of shopping trips.  Each
trip is a tuple $(\mby_t, u_t, w_t, r_t)$ containing a collection of
purchased items ($\mby_t$), the customer who purchased them ($u_t$),
the week that the shopping trip took place ($w_t$), and the prices of
all items ($r_t$).  The likelihood of each shopping trip captures
the sequential process of the customer choosing items
(\Cref{eq:choice-prob}), where each choice is a log linear model
(\Cref{eq:log-linear}).  \shopper\ includes terms for item popularity,
customer preferences, price sensitivity, and seasonal effects
(\Cref{eq:psi_tc}).  We further described a variant that models the
customer thinking ahead and acting optimally
(\Cref{eq:psi_thinkahead}).  Though this is a model of ordered
baskets, we can use it to evaluate the probability of an unordered
basket of items (\Cref{eq:p_y_perm}).

\shopper\ is based on log linear terms that involve several types of
latent parameters: per-item interaction coefficients $\rho_c$,
per-item attributes $\alpha_c$, per-item popularities $\lambda_c$,
per-user preferences $\theta_u$, per-user per-item price sensitivities
$(\gamma_u, \beta_c)$, and per-week per-item seasonal effects
$(\delta_w, \mu_c)$.  Our goal is to estimate these parameters given a
large data set of $T$ trips $\{(\mby_t, u_t, w_t, r_t)\}_{t=1}^{T}$.

We take a Bayesian approach.  We place priors on the parameters to
form a Bayesian model of shopping trips, and then we form estimates of
the latent parameters by approximating the posterior. We use
independent Gaussian priors for the real-valued parameters, $\rho_c$,
$\alpha_c$, $\theta_u$, $\lambda_c$, $\delta_{w}$, and $\mu_c$. We use
gamma priors for the positive-valued parameters associated with price
sensitivity, $\gamma_u$ and $\beta_c$.  With the approximate
posterior, we can use the resulting estimates to identify various
types of purchasing patterns and to make predictions about the
distribution of market baskets; in particular predictions under price
changes.

{Finally, note that }\shopper{ uses matrix factorization approaches for customer
preferences ($\theta_u^\top \alpha_c$), price effects ($\gamma_u^\top \beta_c$),
and seasonal effects ($\delta_w^\top \mu_c$). Matrix factorization has two sources
of invariance (label switching and rescaling). This implies that we cannot
uniquely identify individual variables such as $\theta_u$ or $\gamma_u$ from
the data. Fortunately, this is not an issue for (counterfactual) predictions,
because they only depend on the inner products and not on the individual
coefficients.}




\newcommand{\coffee}{{\it coffee}}
\newcommand{\diapers}{\textit{diapers}}
\newcommand{\candy}{\textit{candy}}
\newcommand{\ramen}{\textit{ramen}}
\newcommand{\hotdogs}{\textit{hot dogs}}
\newcommand{\buns}{\textit{hot dog buns}}
\newcommand{\shells}{\textit{taco shells}}
\newcommand{\seasoning}{\textit{taco seasoning}}

\section{Illustrative simulation}\label{sec:illustrative}
We now describe a simulation of
customers to illustrate how \shopper\ works.  The purpose of our
simulation study is to show \shopper's ability to handle heterogenous
purchases, complements, and counterfactual settings (i.e.,
interventions on price).  More specifically, we ask whether the model
can disentangle correlated preferences and complements, both of which
might contribute to co-purchasing.  We illustrate that distinguishing
between these two types of relationships relies on the thinking ahead
property from \Cref{sec:model_lookahead}.

We simulate the following world.
\begin{itemize}[leftmargin=*]
\item There are $8$ different items: \coffee, \diapers,
  \ramen, \candy, \hotdogs,
  \buns, \shells, and \seasoning.

\item Customers have correlated preferences and there are two types of
  customers: new parents and college students.  New parents frequently
  buy \coffee\ and \diapers; college students frequently buy \ramen\ and
  \candy.  What correlated preferences means is that the preferred
  items (e.g., \coffee\ and \diapers) are decided on independently
  (based on their price).  But note that new parents never buy \ramen\ 
  or \candy, regardless of price, and college students never buy
  \coffee\ or \diapers.

\item The other items represent complementary pairs.  In addition to
  their preferred items, each customer also buys either \hotdogs\ and
  \buns\ or \shells\ and \seasoning.
  In this imaginary world, customers never buy just one
  item in the complementary pair, and they always buy one of the pairs
  (but not both pairs).

\item Customers are sensitive to price.  When the price of a preferred
  item is low (e.g., \coffee\ for a new parent), they buy that item
  with probability $0.95$; when the price of a preferred item is high,
  they buy it with probability $0.1$.  Each customer decides on
  buying their preferred items independently.  That is, a new
  parent makes independent decisions about \coffee\ and \diapers\ (each
  based on their respective prices), and similarly for college students
  and their choices on \candy\ and \ramen.

  Sensitivity to the price of complementary pairs is different,
  because a high price of one of the items in a pair (e.g.,
  \hotdogs) will lower the probability of purchasing the pair
  as a whole.  Specifically, when the price of all complementary pairs
  is low, each customer purchases one or the other with probability
  $0.5$.  When one item (e.g., \hotdogs) has a high price, each
  customer buys the lower priced pair (\shells\ and \seasoning, in this case) with probability
  $0.85$ and buys the pair with the high priced item with probability
  $0.15$.
\end{itemize}
With this specification, we simulate $100$ different customers
($50$ new parents and $50$ college students) and $1000$
trips per customer.  For each trip, the first four items (\coffee,
\diapers, \ramen, \candy) each have a $40\%$ chance of being marked up
to a high price.  Further, there is a $60\%$ chance that one of the
items in the complementary pairs (\hotdogs, \buns,
\shells, \seasoning) is marked up.
At most one of the items in the complementary pairs has a high price.

Given the simulated data, we fit a \shopper\ model that contains terms
for latent attributes, user preferences, and price.  For simplicity,
we omit seasonal effects here.  (The approximate inference algorithm
is described below in \Cref{sec:inference}.) One of our goals is to
confirm the intuition that thinking ahead correctly handles
complementary pairs of items.  Thus we fit \shopper\ both with and
without thinking ahead.

Consider a shopping trip where the prices of \coffee\ and \shells\ are
high, and consider a customer who is a new parent.  As an example,
this customer may buy \diapers, \hotdogs, \buns, and then check
out. \Cref{tab:example_lookahead1} shows the {predicted} probabilities of each
possible item at each stage of the shopping trip. They are illustrated
for the models with and without thinking ahead.

First consider the preference items in the first stage, before
anything is placed in the basket. The customer has a high probability
of purchasing \diapers, and low probabilities of purchasing \coffee,
\ramen, and \candy.  The probabilities for \ramen\ and \candy\ are low
because of the type of customer (new parent); the probability for \coffee\ is low
because of its high price.

Now consider the complementary pairs.  In the model without thinking
ahead, the customer has high probability of buying \hotdogs, \buns,
and \seasoning; because of its high price, she has a low probability of
buying \shells.  But this is incorrect.  Knowing that the price of
\shells\ is high, she should have a lower probability of buying
\seasoning\ because it is only useful to buy along with \shells.  The
thinking-ahead model captures this, giving both \seasoning\ and
\shells\ a low probability.

Subsequent stages further illustrate the intuitions behind \shopper.
First, each stage zeros out the items that are already in the basket
(e.g., at stage 2, \diapers\ have probability $0$).
Second, once one item of the complementary pair is bought, the probability
of the other half of the pair increases and the probabilities of the
alternative pair becomes low.  In this case, once the customer buys
\hotdogs, the probability of the taco products goes to zero and the
probability of \buns\ increases.


\begin{table}[t]
	\centering
	\scriptsize
	\begin{tabular}{clllll}\toprule
		& & stage 1: \diapers & stage 2: \hotdogs & stage 3: \buns & stage 4: \textit{checkout}\\ \midrule
\parbox[t]{2mm}{\multirow{9}{*}{\rotatebox[origin=c]{90}{non think-ahead}}} & diapers & {\color{myGray} $\mathbf{0.31}$} & {\color{myGray} $0.00$} & {\color{myGray} $0.00$} & {\color{myGray} $0.00$} \\ 
 & coffee ($\uparrow$) & {\color{myGray} $0.03$} & {\color{myGray} $0.02$} & {\color{myGray} $0.05$} & {\color{myGray} $0.21$} \\ 
 & ramen & {\color{myGray} $0.00$} & {\color{myGray} $0.00$} & {\color{myGray} $0.00$} & {\color{myGray} $0.00$} \\ 
 & candy & {\color{myGray} $0.00$} & {\color{myGray} $0.00$} & {\color{myGray} $0.00$} & {\color{myGray} $0.00$} \\ 
 & hot dogs & {\color{myGray} $0.18$} & {\color{myGray} $\mathbf{0.25}$} & {\color{myGray} $0.00$} & {\color{myGray} $0.00$} \\ 
 & hot dog buns & {\color{myGray} $0.17$} & {\color{myGray} $0.25$} & {\color{myGray} $\mathbf{0.79}$} & {\color{myGray} $0.00$} \\ 
 & taco shells ($\uparrow$) & {\color{myGray} $0.14$} & {\color{myGray} $0.19$} & {\color{myGray} $0.00$} & {\color{myGray} $0.00$} \\ 
 & taco seasoning & {\color{BrickRed} $0.17$} & {\color{BrickRed} $0.24$} & {\color{myGray} $0.00$} & {\color{myGray} $0.00$} \\ 
 & checkout & {\color{myGray} $0.00$} & {\color{myGray} $0.05$} & {\color{myGray} $0.16$} & {\color{myGray} $\mathbf{0.79}$} \\ \midrule
\parbox[t]{2mm}{\multirow{9}{*}{\rotatebox[origin=c]{90}{think-ahead}}} & diapers & {\color{myGray} $\mathbf{0.37}$} & {\color{myGray} $0.00$} & {\color{myGray} $0.00$} & {\color{myGray} $0.00$} \\ 
 & coffee ($\uparrow$) & {\color{myGray} $0.02$} & {\color{myGray} $0.02$} & {\color{myGray} $0.07$} & {\color{myGray} $0.10$} \\ 
 & ramen & {\color{myGray} $0.00$} & {\color{myGray} $0.00$} & {\color{myGray} $0.00$} & {\color{myGray} $0.00$} \\ 
 & candy & {\color{myGray} $0.00$} & {\color{myGray} $0.00$} & {\color{myGray} $0.00$} & {\color{myGray} $0.00$} \\ 
 & hot dogs & {\color{myGray} $0.24$} & {\color{myGray} $\mathbf{0.34}$} & {\color{myGray} $0.00$} & {\color{myGray} $0.00$} \\ 
 & hot dog buns & {\color{myGray} $0.24$} & {\color{myGray} $0.42$} & {\color{myGray} $\mathbf{0.85}$} & {\color{myGray} $0.00$} \\ 
 & taco shells ($\uparrow$) & {\color{myGray} $0.06$} & {\color{myGray} $0.10$} & {\color{myGray} $0.00$} & {\color{myGray} $0.00$} \\ 
 & taco seasoning & {\color{ForestGreen} $0.06$} & {\color{ForestGreen} $0.10$} & {\color{myGray} $0.00$} & {\color{myGray} $0.00$} \\ 
 & checkout & {\color{myGray} $0.00$} & {\color{myGray} $0.02$} & {\color{myGray} $0.08$} & {\color{myGray} $\mathbf{0.90}$} \\ \bottomrule
	\end{tabular}
	\caption{An example of the predicted probabilities for a new parent customer in the toy simulation, for a basket that contains \diapers, \hotdogs, and \buns. \textbf{(Top)} Model without the thinking-ahead property. \textbf{(Bottom)} Model with the thinking-ahead property. In the table, each column represents a step in the sequential process, and the numbers denote the probability of purchasing each item. We have marked with bold font the items purchased at each step. The arrows pointing up for \coffee\ and \shells\ indicate that these items have high price in this shopping trip. The thinking-ahead model provides higher predictive log-likelihood, because it can capture the joint distribution of pairs of items: it correctly assigns lower probability to \seasoning\ at each step, because \shells\ have high price.\label{tab:example_lookahead1}}
\end{table}


As a final demonstration on simulated data, we generate a test set
from the simulator with $30$ shopping trips for each customer. On this
test set, we ``intervene'' on the price distribution: the probability
of a preference item having a high price is $0.95$ and one of the four
complementary items always has a high price.
On this test set, a better model will provide higher held-out log
probability and we confirm that thinking ahead helps. The
thinking-ahead model gives an average held out log probability of
$-2.26$; the model without thinking ahead gives an average held out
log probability of $-2.79$.


\section{Related work}
\label{sec:related}

\shopper\ relates closely to several lines of research in machine
learning and economics.  We discuss them in turn.

\subsection{Machine learning: Word embeddings and recommendation}

One of the central ideas in \shopper\ is that the items in the store
have latent vector representations and we estimate these
representations from observed shopping basket data.  In developing
\shopper, we were directly inspired by the neural probabilistic
language model of \citet{Bengio2003,bengio2006neural}.  That model
specifies a joint probability of sequences of words, parameterized by
a vector representation of the vocabulary.  In language, vector
representations of words (also called ``distributed representations'')
allow us to reason about their usage and meaning
\citep{Harris1954,Firth1957,Bengio2003,Mikolov2013distributed}.
Here we expand on this idea to
study consumer behavior, and we show how the same mathematical
concepts that inspired word representations can help understand
large-scale consumer behavior.

We note that neural language models have spawned myriad developments
in other so-called word embedding methods for capturing latent
semantic structure in language
\citep{Mnih2007,Mnih2012,Mikolov2013efficient,Mikolov2013distributed,
Mikolov2013linguistic,Mnih2013learning,Pennington2014,Levy2014neural,
Vilnis2014,Arora2016,Barkan2016bayesian,Bamler2017}.
There has been some work on extrapolating embedding ideas to shopping
items \citep{Rudolph2016,Liang2016,Barkan2016}, but these newer
methods are not readily appropriate to the analyses here.  The reason
is that they are conditionally specified models, defined by the
conditional distribution of each item given the others.  In contrast,
\shopper\ directly models the joint distribution of items; this
enables \shopper\ to more easily account for additional information,
such as price and customer preferences.  From the econometric
perspective, casting \shopper\ as a proper generative model enables a
direct interpretation as a structural equation model with meaningful
parameters.  This lets us use \shopper\ to make counterfactual
predictions, such as about how price changes will affect firm revenue
and consumer welfare.

The way \shopper\ models customer preferences is similar to modern
recommender systems, where matrix factorization models are standard.
Matrix factorization decomposes observed data of customer/item
interactions into latent preferences and item attributes
\citep{Canny2004,Hu2008,Ma2011,Wang2011,Gopalan2015},
{potentially with an infinite number of latent
parameters }\citep{Gorur2006,Gopalan2014}{. Matrix factorization
methods have been used for web services
recommendations }\citep{Stern2009}{ and also market basket
data }\citep{Wan2017}. Though it is effective, matrix
factorization does not directly capture
item-to-item co-occurrence, which is the main motivation behind models
based on latent attributes.  \shopper\ includes both attributes and
preferences, thereby combining ideas from distributed representations
with those from recommendation.  \shopper\ also goes further by
including price (and price sensitivities) and seasonal effects, and
generally seeking realistic model of consumer behavior.  It is more
appropriate than machine learning recommendation methods for
evaluating counterfactual questions.
{Incorporating observed attributes also relates to
recommender systems with
attributes }\citep[see, e.g.,][]{Abernethy2009}{, although }\shopper{ can
consider time-varying item attributes such as prices.}



\subsection{Economics and marketing: Discrete choice and utility
  maximization}

\shopper\ expands on discrete choice models of consumer behavior,
models widely used in economics and marketing.  The majority of papers
in this literature analyze the discrete choice of a consumer who
selects a single product from a set of prespecified imperfect
substitutes; examples include choices of laundry detergent, personal
computers, or cars.  As reviewed by \citet{ keane2013panel}, this
literature focuses on estimating cross-price elasticities, accounting
for (latent) consumer heterogeneity in tastes. A small literature on
``market mapping''
\citep{elrod1988choice,elrod1995factor,chintagunta1994heterogeneous}
considers latent attributes of items within a product category (e.g.,
laundry detergent).  \shopper\ is similar in spirit in its focus on
estimating latent characteristics of items using panel data.  However,
\shopper\ differs in the scale it handles, considering hundreds or
thousands of items and hundreds of thousands of shopping trips.  This
is not simply a matter of improving computational speed; analyzing
the whole market for counterfactual prediction requires a more
thorough treatment of issues like complementarity, substitutability,
and customer heterogeneity.  
{A complementary approach to ours is the work
by }\citet{Semenova2018}{, who consider observational
high-dimensional product attributes (e.g., text descriptions and images)
rather than latent features.} Also see our related paper for another
approach at this same scale~\citep{athey2017counterfactual}.

\shopper\ also differs significantly from other discrete choice models
in that it considers interaction effects among a large number of items
in a consumer's basket without a priori imposing a structure of
product relationships.  A variety of datasets track a panel of
consumers over time, recording their baskets in each shopping trip.
This includes supermarket shopping datasets and online panel datasets
of shopping from vendors like Neilsen and comScore, as well as data
collected internally by online and offline retailers.  However, there
is relatively little work in economics and marketing that analyzes
consumers' choices over many items in a basket using this type of
data.  Even when multiple items are considered, few methods attempt to
estimate directly whether items are substitutes or complements;
instead, products are typically assumed to be independent of one
another across categories, or strong substitutes within categories (in
the sense that purchasing one item precludes purchasing
others). \shopper\ does not require a prespecified ontology of
categories; the interaction effects are estimated directly from
the data.



Of course, estimating the extent to which products are substitutes and
complements for one another is a hard problem when there are many
products: if there are $C$ products, there are $2^C$ baskets, and
without further restrictions, a consumer's preferences over baskets
thus has $2^C$ parameters.  One approach is to make use of more
aggregate data combined with functional form assumptions; for example,
the almost ideal demand system \citep{deaton1980almost} considers a
consumer maximizing utility subject to a budget constraint over a
period of time, where the share of budget allocated to each good is
taken to be continuous.  With functional form assumptions, the
consumer's expenditure share for each product can be written as a
linear function of transformations of the prices of other products,
and the parameters of these demand functions can be interpreted as
expressing underlying consumer preferences, including the extent to
which products are substitutes or complements.  This approach has a
number of attractive features, including tractability and
computational simplicity, and it can be used to handle many products
simultaneously.  However, in reality prices often change relatively
frequently, and without data about when a consumer shopped or what
prices the consumer saw, inferences drawn from aggregate data can be
misleading.  In addition, more aggregated data loses crucial
information about the co-purchases an individual makes on the same
trip.

A small set of papers in economics and marketing \citep{athey1998empirical}
attempts to use individual choice
data estimate parameters that describe interaction effects
(substitutes or complements) with a very small number of items; see
\citet{chintagunta2011structural} and \citet{berry2014structural} for
recent surveys of the literature.  For example,
\citet{train1987demand} treat each bundle as a discrete alternative,
but use nested logit to account for correlation among related
bundles, while \citet{gentzkow2007valuing} incorporates a parameter in
the utility for complementarity among two items.
\citet{song2007discrete} build a utility-maximization framework where
consumers select not just whether to purchase, but how much, and
apply it to supermarket purchase data for two products, laundry detergent and
fabric softener.  However, due to computational considerations, most of the
papers that deal with disaggregated data have focused on a very small
number of products and a small number of customers. With \shopper, we are
able to jointly model thousands of items and millions of purchased items.





\section{Inference}
\label{sec:inference}
\Cref{sec:model} defined \shopper, a Bayesian model of market baskets.
Given a large data set, our goal is to infer the latent parameters,
the components that govern item attributes, user preferences, price
sensitivity, and seasonal effects.  We then use the inferred parameters
to make predictions, e.g., about demand, and to characterize patterns of
purchase behavior. In this section we describe a
variational inference algorithm to solve this problem.  

For conciseness, we denote the collection of latent variables as
$\mb{\ell} = \{\rho,\alpha, \lambda, \theta, \gamma, \beta,
\mu, \delta\}$, the observed shopping baskets as
$\mathpzc{y}=\{\mathpzc{y}_t\}$, and the observed characteristics of
shopping trips as $\bx = x_{1:T}$, where $x_t = (u_t, w_t, r_t)$
indicates the customer, calendar week, and item prices for the $t$th trip.
The posterior is
\begin{align}
  p(\mb{\ell} \g \mathpzc{y}, \bx) = \frac{p(\mb{\ell}) \prod_{t=1}^{T} p(\mathpzc{y}_t \g
  \mb{\ell},x_t)}{p(\mathpzc{y} \g \bx)}.
\end{align}
It is difficult to compute the posterior in closed form because of the intractable
denominator, the marginal likelihood of the observed data. Further, as
we described in \Cref{sec:model}, the likelihood term is itself
intractable because of the sum over the large number of items in the
denominator of the softmax function.  We must use approximate Bayesian
inference.

Variational inference~\citep{Jordan1999,Wainwright2008} is an
alternative to MCMC for approximating otherwise intractable Bayesian
posteriors.  Compared to MCMC, it more easily scales to large data,
and especially so in non-conjugate models such as \shopper.

Variational inference approximates the posterior with a parameterized
family of distributions of the latent variables $q(\mb{\ell} ; \nu)$;
the parameters $\nu$ are called variational parameters.  The idea is
to find the member of the family that is closest to the exact
posterior, where closeness is measured by the \gls{KL} divergence.
Minimizing the \gls{KL} divergence is equivalent to maximizing the \gls{ELBO},
\begin{align}\label{eq:elbo}
  \Lcal(\nu) = \E{q(\mb{\ell} ; \nu)}{\log p(\mb{\ell}, \mathpzc{y} \g \bx) - \log q(\mb{\ell} ; \nu)},
\end{align}
where the expectations are taken with respect to the variational
distribution.

The \gls{ELBO} $\Lcal(\nu)$ provides a lower bound on the marginal
likelihood $\log p(\mathpzc{y} \g \bx)$, and hence its name.  Its maximizer is
$\nu^* = \arg \max_{\nu} \Lcal(\nu)$ (which also minimizes the
\gls{KL} divergence).  When we use variational inference, we first optimize the
\gls{ELBO} and then use the resulting distribution
$q(\mb{\ell} ; \nu^*)$ as a proxy for the exact posterior, e.g., to
investigate the inferred hidden structure or to approximate the
posterior predictive distribution.  In summary, variational inference
uses optimization to approximate the posterior.  For a review of
variational inference, see \citet{Blei:2016}.

To develop a variational inference algorithm for \shopper, we first
specify the variational family and then develop the optimization
procedure for fitting its parameters.  \Cref{sec:supp_inference}
\citep{Ruiz2017_supp} gives all the details; here we sketch the main
ideas.

The first step to deriving a variational inference algorithm is to
specify the variational family $q(\mb{\ell} ; \nu)$.  Following many
successful applications of variational inference, we posit the
mean-field family, where each latent variable is governed by its own
distribution and endowed with its own variational parameter.  We use
Gaussian variational factors for the latent variables with Gaussian
priors; and gamma variational factors for the latent variables with
gamma priors. Though the mean-field variational family makes strong
independence assumptions, we emphasize that latent variables are not
identically distributed and thus the variational family is still very flexible.
Thus, for example, the variational distribution of the latent
attributes of peanut butter will peak at one point in attribute space;
the latent attributes of taco shells will (likely) peak at a
different point.

Given the variational family, we next formulate how to optimize
\Cref{eq:elbo} with respect to the variational parameters; we use
gradient-based stochastic optimization.  This problem is complicated for several
reasons: the data set sizes are large, the expectations are
intractable, and the likelihood is expensive to calculate.  To
overcome these obstacles, we bring together a variety of recent
innovations around variational inference, summarized below and
detailed in the supplement \citep{Ruiz2017_supp}.

\begin{itemize}[leftmargin=*]

\item The first issue is that the data sets are large.
  The gradient contains a term for every item purchased
  (e.g., around $6$ million terms for the experiments of
  \Cref{sec:experiments}),
  and this is too expensive to be practical,
  especially because we need to calculate it at each iteration.
  Thus we lean on stochastic
  optimization~\citep{Robbins1951,Blum1954,Bottou:2016}, where we follow
  cheaper-to-compute unbiased noisy estimates of the
  gradient.  Following \citet{Hoffman2013}, we calculate noisy
  gradients by subsampling from the data and taking a scaled gradient
  relative to the subsample.

\item The second issue is that, even setting aside the large data, the
  expectations in \Cref{eq:elbo} are analytically intractable.  
  Again we sidestep this issue with
  stochastic optimization, devising the gradient itself as an
  expectation and then forming unbiased noisy gradients with Monte
  Carlo approximations.  In particular, we use the reparameterization
  gradient \citep{Kingma2014,Titsias2014_doubly,Rezende2014} and its
  generalizations \citep{Ruiz2016nips,Naesseth2017}.

\item The third issue is that we need to evaluate the probability of
  unordered baskets (\Cref{eq:p_y_perm}), which involves an expensive
  summation over permutations of items, even for moderately-sized
  baskets. {We address this by deriving a variational bound of the
  }\gls{ELBO}{, $\Lcal(\nu)\geq \widetilde{\Lcal}(\nu)$, and using
    stochastic optimization to maximize $\widetilde{\Lcal}(\nu)$ with
    respect to the variational parameters.  We obtain the bound
    $\widetilde{\Lcal}(\nu)$ by introducing an auxiliary variational
    distribution over the permutations $\pi$, which we set equal to
    the prior (i.e., uniform), and applying Jensen's inequality. This
    step} resembles the variational inference procedure for the Indian
  buffet process \citep{DoshiVelez2009}.  {Crucially, the bound
    $\widetilde{\Lcal}(\nu)$ contains a direct summation over
    permutations. To lower the computational cost, we use stochastic
    variational inference }\citep{Hoffman2013}{. Specifically, we
    randomly samples one (or a few) permutations $\pi$ for each basket
    at each iteration of variational inference.  This provides
    unbiased estimates of the gradient of the bound
    $\widetilde{\Lcal}(\nu)$, which addresses the computational
    complexity of }\Cref{eq:p_y_perm}{, though at the expense of
    having introduced an additional variational bound. Our empirical results
  indicate that this is an effective procedure for addressing the
  unknown permutation of the items.}

\item The final computational issue is how to evaluate likelihoods of
  the form in \Cref{eq:choice-prob}.  This is expensive because of the
  number of items to sum over in the denominator.  We solve the
  problem with the approach developed by
  \citet{Titsias2016}. This approach derives a bound for which we can obtain
  cheap unbiased estimates of the gradient. (Alternatively, we
    could adapt the approach of \citet{Ruiz2018}.)
\end{itemize}

See \url{https://github.com/franrruiz/shopper-src} for a publicly
available implementation of \shopper.

\section{Empirical Study}
\label{sec:experiments}

We now use \shopper\ to study a real-world data set of consumers.  Our
dataset contains $570{,}878$ baskets from an large grocery store
(anonymized for data privacy reasons) \citep{Che2012}.
The dataset is available to researchers at Stanford and Berkeley
by application; it has been used previously in other research papers
(see \url{https://are.berkeley.edu/SGDC}).
These baskets contain $5{,}968{,}528$ purchases of $5{,}590$ unique items.  The data
spans $97$ weeks and $3{,}206$ different customers. We split the baskets
into {a training, validation, and test set. The test set contains
all the baskets in the last $2$ months of the data collection period; the
validation set contains $5\%$ of the remaining purchases, chosen at random.}

We study the predictive performance of \shopper\, both observationally
and under price intervention.  Note that the store changes prices regularly,
and this variation allows us to empirically separate correlated latent
preferences from price effects, once we control for seasonality.

We also demonstrate how to use \shopper\ to qualitatively analyze the data.
It finds interpretable latent features, which can be used to find complementary and
exchangeable items.

We first focus on category-level data, where we group individual
items---according to their \gls{UPC}---into
their category (such as ``oranges,'' ``flour,'' ``pet supplies,'' 
etc.); there are $374$ unique categories.
The category-level data contains fewer items and is computationally
less demanding for any model; so we use these data to compare the
predictive performance of \shopper\ to other models.  We obtain better
predictive accuracy (in terms of test log-likelihood) than the
competing methods, especially under price interventions.

Secondly, we focus on \gls{UPC} data, all $5{,}590$ items.  Based only
on posterior inference with the observed basket data, \shopper\
identifies characteristics of items, including complements and
exchangeable items. We define quantities to
measure these concepts, and we find that
exchangeability is a good proxy to find substitutes.



\subsection{Category-level data}
\label{sec:experiments_category}

We compare \shopper\ to other latent factor models of shopping data.
In particular, we apply \gls{HPF} \citep{Gopalan2015} and exponential
family embeddings \citep{Rudolph2016}. \gls{HPF} focuses on user
preferences; exponential family embeddings focuses on item-to-item
interactions. \Cref{tab:model_comparison} summarizes the differences
between \shopper\ and these models; note that none of the comparison
models captures price or seasonal effects.

{We also fitted two simpler models. First, we studied
a multinomial logistic regression model that predicts an item conditioned on the
rest of the items in the basket and the item prices. The inputs are binary indicators
of the items in the basket excluding the target item, as well as the logarithm
of the normalized prices of all the items. Second, we studied factor
analysis }\citep{Cattell1952}{ on the basket-by-item matrix.
However, we do not report the results of these two models in the paper because they gave
worse predictions than all other methods on the test set, corresponding to the last two
months of data.} (On the validation set, the multinomial logistic
regression model performed slightly better than exponential family embeddings but worse
than \gls{HPF}. We emphasize that multinomial logistic regression takes the item
prices as inputs, in contrast to exponential family embeddings and \gls{HPF}.)

\begin{table}[t]
  \centering
  \scriptsize
  \begin{tabular}{cccccc}\toprule
    \multirow{2}{*}{Model} & \multirow{2}{*}{Data} & User        & Item-to-item & \multirow{2}{*}{Price} & Seasonal \\
                           &                       & preferences & interactions &                        &  effects \\ \midrule
    B-Emb \citep{Rudolph2016} & Binary & $\times$ & \checkmark & $\times$ & $\times$ \\
    P-Emb \citep{Rudolph2016} & Count & $\times$ & \checkmark & $\times$ & $\times$ \\
    \acrshort{HPF} \citep{Gopalan2015} & Count & \checkmark & $\times$ & $\times$ & $\times$ \\
    \shopper\ (this paper) & Binary & \checkmark & \checkmark & \checkmark & \checkmark \\ \bottomrule
  \end{tabular}
  \caption{We compare properties of \shopper\ to existing models of consumer
    behavior.   Bernoulli embeddings (B-Emb) and Poisson embeddings
    (P-Emb) only model in-basket item-to-item interactions; \gls{HPF}
    factorizes the user/item matrix.  \shopper\ models both types of
    regularity and additionally adjusts for price and seasonal effects.%
    \label{tab:model_comparison}}
\end{table}

\subsubsection{Quantitative results}
We fit each model to the category-level data.  In \shopper\,
we set most of the Gaussian hyperparameters to
zero mean and unit variance.  (The seasonal effect hyperparameters
have a smaller variance, 0.01, because we do not expect large seasonal
effects.)  The price sensitivity parameters have Gamma priors; we use
a shape of $1$ and a rate of $10$.  As for the comparison models, we
modify \gls{HPF} to allow for multiple shopping trips of the same
user; in its original construction it can only capture a single
trip. Finally, we implement two versions of exponential family
embeddings, Bernoulli embeddings for binary data (labeled ``B-Emb'')
and Poisson embeddings for count data (labeled ``P-Emb'').  We weight
the zeros by a factor of $0.1$, as suggested by
\citet{Rudolph2016}. For all methods, we use the validation set to
assess convergence.


{To choose the number of latent factors $K$, we first set the number
of factors of the price and seasonal vectors to $10$ and run }\shopper{ with
$K\in\{10,20,50,100,200\}$. We choose the value $K=100$ because it provides
the best predictions on the validation set. For that value of $K$, we then 
explore the number of latent factors for the price and seasonal vectors in the
set $\{5,10,20,50\}$, finally choosing $10$. We also set $K=100$
for }\gls{HPF}{ and exponential family embeddings.}

To evaluate the models, we calculate the average log-likelihood of the
test-set items.  For each, we calculate its probability conditioned
on the other observed items in the basket.  Higher log probability
indicates a better model fit.  \Cref{tab:test_llh} shows the results.
{The numbers in parentheses show the standard deviation, obtained by
using bootstrap on the test samples.}

We study several types of test sets.  The second column corresponds to
a typical test set, {containing two months of data}.  Columns 3-5
focus on skewed test sets, {where the target items have more extreme
prices with respect to their average price within each month (outside of the range
$\pm 2.5\%$, $\pm 5\%$, and $\pm 15\%$). To control for seasonal effects,
we consider items whose price is outside that range with respect to
the per-month average price.}
These evaluations are
suggestive of the performance under price intervention, i.e., where
the distribution of price is different in testing than it is in
training. The numbers in parentheses {on the table heading} indicate the
number of purchases considered in each column.

We report results that incrementally add terms to the basic \shopper\
model.  The most basic model contains user information with
item-to-item interactions (``I+U''); it improves predictions over the
competing models.  {The next model includes seasonal effects
(``I+U+S''); it improves performance on the typical test set
only.} (To compute predictions for the models with seasonal
effects, we set the seasonal parameter $\delta_w$ for the weeks in
the test set equal to the value of $\delta_w$ corresponding to the
same week but the year before.)
{We also consider a model that} includes price sensitivity
(``I+U+P''); it further improves performance.  The full model adds
seasonal effects (``I+U+P+S''); {in general} it gives the best
performance.  As expected, modeling price elasticity is important in making
counterfactual predictions.  The performance gap between the models
with and without price increases as the test-set prices diverge from
their average values.

\Cref{tab:test_llh}{ compares models with different numbers of
  latent parameters. (}\Cref{tab:number_params}{ gives the number
  of latent parameters in each model.) Exponential family
  embeddings posit two $K$-length vectors for each
  item. }\gls{HPF}{ has one $K$-length vector for each user and
  item. }\shopper{ has one $K$-length vector for each user and two
  $K$-length vectors for each item, in addition to the intercept terms
  and the price and seasonal components.}

\begin{table}[t]
  \centering
  \scriptsize
  \begin{tabular}{ccccc}\toprule
    \multirow{3}{*}{Model} & \multicolumn{4}{c}{Log-likelihood} \\
                           & All & Price$\pm2.5\%$ & Price$\pm5\%$ & Price$\pm15\%$ \\
                           & ($540$K) & ($231$K) & ($139$K) & ($25$K) \\ \midrule
    B-Emb \citep{Rudolph2016} & $-5.119$ ($0.001$) & $-5.119$ ($0.002$) & $-5.148$ ($0.002$) & $-5.250$ ($0.006$) \\
    P-Emb \citep{Rudolph2016} & $-5.160$ ($0.001$) & $-5.138$ ($0.002$) & $-5.204$ ($0.002$) & $-5.311$ ($0.005$) \\
    \acrshort{HPF} \citep{Gopalan2015} & $-4.914$ ($0.002$) & $-4.931$ ($0.002$) & $-4.994$ ($0.003$) & $-5.061$ ($0.009$) \\
    \shopper\ (I+U) & $-4.744$ ($0.002$) & $-4.743$ ($0.003$) & $-4.776$ ($0.003$) & $-4.82$ ($0.01$) \\
    \shopper\ (I+U+S) & $-4.730$ ($0.002$) & $-4.778$ ($0.003$) & $-4.801$ ($0.004$) & $-4.83$ ($0.01$) \\
    \shopper\ (I+U+P) & $-4.728$ ($0.002$) & $-4.753$ ($0.003$) & $\mathbf{-4.747}$ ($0.004$) & $-4.69$ ($0.01$) \\
    \shopper\ (I+U+P+S) & $\mathbf{-4.724}$ ($0.002$) & $\mathbf{-4.741}$ ($0.003$) & $-4.774$ ($0.004$) & $\mathbf{-4.64}$ ($0.01$) \\ \bottomrule
  \end{tabular}
  \caption{{Average predictive log-likelihood on the test set,
    conditioning on the remaining items of each basket
    (the numbers in parentheses indicate standard deviation).
    }\shopper{ with user preferences improves over the existing models.
    The improvement grows when adjusting for price
    and seasonal effects, and especially so when using skewed test
    sets that emulate price intervention.}  \label{tab:test_llh}}
\end{table}

\begin{table}[t]
  \centering
  \scriptsize
  \begin{tabular}{ccccc}\toprule
    \multirow{2}{*}{Model} & \multicolumn{4}{c}{Number of latent parameters} \\
                           & per user & per item & per week & total \\ \midrule
    B-Emb \citep{Rudolph2016} & $-$ & $200$ & $-$ & $74{,}800$ \\
    P-Emb \citep{Rudolph2016} & $-$ & $200$ & $-$ & $74{,}800$ \\
    \acrshort{HPF} \citep{Gopalan2015} & $100$ & $100$ & $-$ & $358{,}000$ \\
    \shopper\ (I+U) & $100$ & $201$ & $-$ & $395{,}975$ \\
    \shopper\ (I+U+S) & $100$ & $211$ & $10$ & $400{,}615$ \\
    \shopper\ (I+U+P) & $110$ & $211$ & $-$ & $431{,}785$ \\
    \shopper\ (I+U+P+S) & $110$ & $221$ & $10$ & $432{,}685$ \\ \bottomrule
  \end{tabular}
  \caption{{Number of latent parameters for each of the considered models.}  \label{tab:number_params}}
\end{table}

\begin{table}[t]
	\centering
	\scriptsize
	\begin{tabular}{ccc}\toprule
		& Three items & Entire baskets \\ \midrule
		Non think-ahead & $-4.795$ ($0.005$) & $-4.96$ ($0.02$) \\
		Think-ahead & $\mathbf{-4.747}$ ($0.004$) & $\mathbf{-4.91}$ ($0.02$) \\ \bottomrule
	\end{tabular}
	\caption{Average test log-likelihood per item, for the models with and without the thinking-ahead property. \textbf{(Left)} Per-item average log-likelihood computed for triplets of items in each basket of size at least 3. \textbf{(Right)} Per-item average log-likelihood on entire held-out baskets.
	{The numbers in parentheses indicate standard deviation of bootstrapped test baskets.}%
	\label{tab:test_llh_baskets}}
\end{table}

Finally, we study the empirical performance of ``thinking ahead,''
particularly when predicting groups of items.
\Cref{tab:test_llh_baskets} shows two metrics: the first column is the
average (per-item) test log-likelihood over three items, conditioned
on the rest of items in the basket; the second column is average
(per-item) log-likelihood over the entire basket.  (Here we exclude
the checkout item{, and we compute the predictions based on the
ordering in which items are listed in the test set}.)
The model with ``thinking ahead'' more correctly handles complements
and price sensitivity, and it provides better predictive performance.

\subsubsection{Qualitative results} \shopper\ provides a better
predictive model of consumer behavior.  We now use the fitted model to
qualitatively understand the data.

\begin{figure}[t]
	\subfloat[{Pet food and supplies.}]{\includegraphics[width=0.44\textwidth]{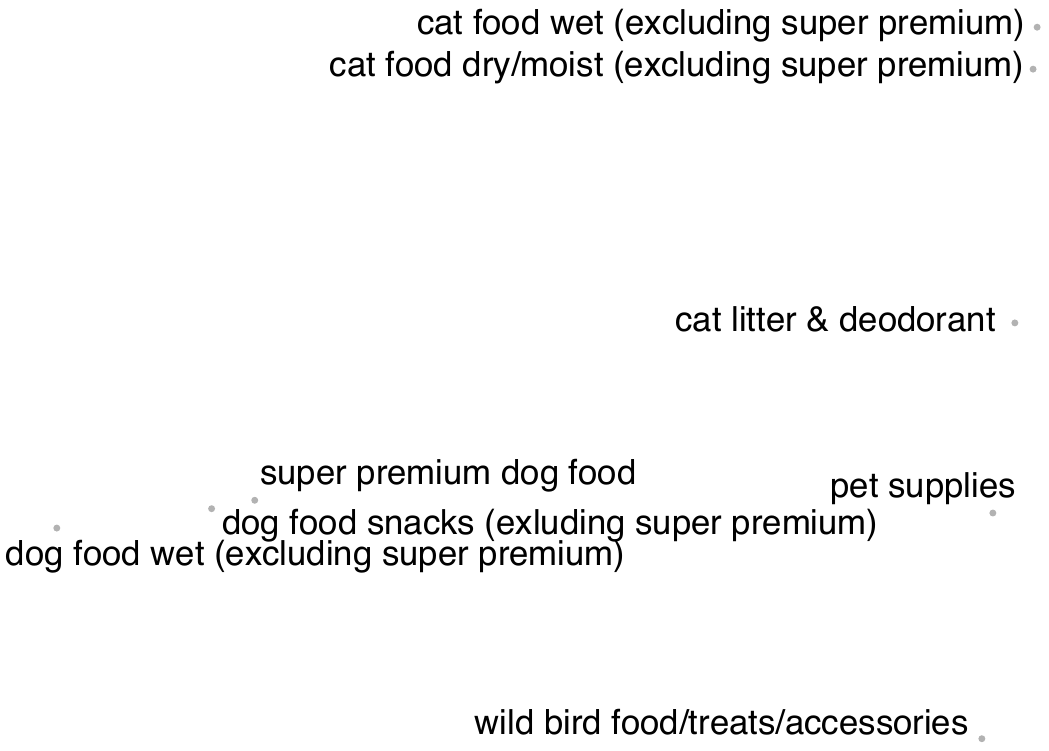}} \hfill
	\subfloat[{Cleaning and hygiene.}]{\includegraphics[width=0.445\textwidth]{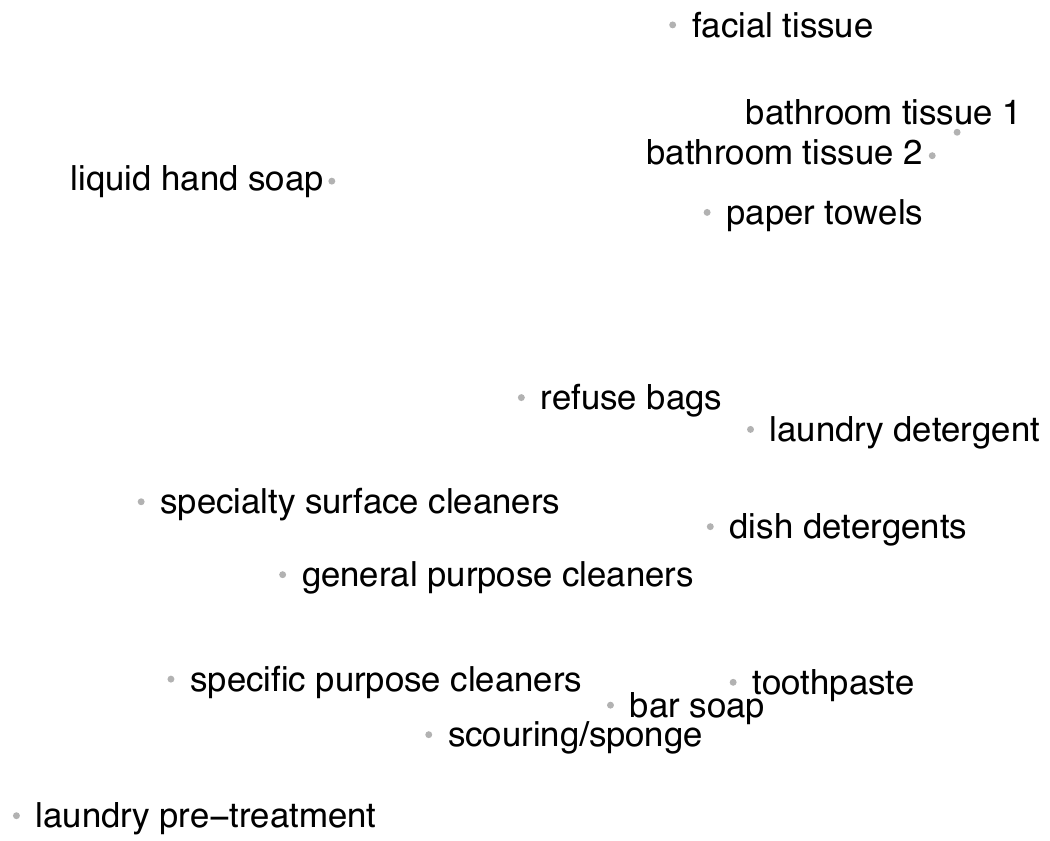}}
	\caption{{Two regions of the two-dimensional }\gls{tSNE}{ projection of the features vectors $\alpha_c$ for the category-level experiment.}\label{fig:tsne_group}}
\end{figure}

First, we assess the attributes vectors $\alpha_c$, confirming that
they capture meaningful dimensions of the items. (Recall that each is
a 100-dimensional real-valued vector.)  As one demonstration, we
project them on to 2-dimensional space using \gls{tSNE}
\citep{vanderMaaten2008}, and
then examine the items in different regions of the projected space.
\Cref{fig:tsne_group} shows two particular regions: one collects
different types {of pet food and supplies}; the other collects different
cleaning products.  As a second demonstration, we can use the cosine distance
to find similar items similar to a ``query item.''
\Cref{tab:mostsimilar_group} shows the top-three most similar items to
a set of queries.


\begin{table}[h]
	\scriptsize
	\begin{tabular}{cccc} \toprule
		mollusks & organic vegetables & granulated sugar & cat food dry/moist \\ \midrule
		finfish all other - frozen & organic fruits & flour & cat food wet \\
		crustacean non-shrimp & citrus & baking ingredients & cat litter \& deodorant \\
		shrimp family & cooking vegetables & brown sugar & pet supplies \\ \bottomrule
	\end{tabular}
	\caption{{Most similar items to a query item (given in the first row), according to the cosine distance between the item features $\alpha_c$, for the category-level experiment.}}
	\label{tab:mostsimilar_group}
\end{table}

\begin{table}[t]
	\scriptsize
	\begin{tabular}{rlrlrl} \toprule
		\multicolumn{2}{c}{Halloween candy} & \multicolumn{2}{c}{cherries} & \multicolumn{2}{c}{turkey - frozen} \\ \midrule
		$3.46$ &	$2006/10/25$ & $3.07$ &	$2006/06/28$ & $3.56$ &	$2005/11/16$ \\
		$3.34$ &	$2005/10/26$ & $3.01$ &	$2006/07/12$ & $3.30$ &	$2006/11/15$ \\
		$2.81$ &	$2005/10/19$ & $2.85$ &	$2006/06/21$ & $2.64$ &	$2005/11/23$  \\
		 &	 &  &	  \multicolumn{1}{l}{$\vdots$} &	&	\\
		$-1.28$ &	$2005/11/23$ & $-3.59$ &	$2006/10/11$ & $-1.25$ &	$2006/06/21$ \\
		$-1.31$ &	$2007/01/03$ & $-3.89$ &	$2006/10/18$ & $-1.29$ &	$2006/07/05$ \\
		$-1.33$ &	$2005/11/16$ & $-4.54$ &	$2006/10/25$ & $-1.30$ &	$2006/07/19$ \\ \bottomrule
	\end{tabular}
	\caption{{Highest and lowest seasonal effects, as given by $\mu_c^\top\delta_w$, for three example items. The model finds the effects of holidays such as Halloween or Thanksgiving, as well as the seasonal availability of fruits.}}
	\label{tab:weekeffects_group}
\end{table}

Other latent variables reveal different aspects of consumer behavior.
For example, \Cref{tab:weekeffects_group} show the highest and lowest
seasonal effects for a set of items. The model correctly captures how
Haloween candy is more popular near Halloween; and turkey is more
popular near Thanksgiving.  It also captures the seasonal availability
of fruits, e.g., cherries.

These investigations are on the category-level analysis.  For more
fine-grained qualitative assessments---especially those around
complementarity and exchangeability---we now turn to the \gls{UPC}-level
model.

\subsection{UPC-level data}
\label{sec:experiments_upc}

We fit \shopper\ to \gls{UPC}-level data, which contains $5$,$590$
unique items.  {We use the same dimensionality of the latent vectors
as in }\Cref{sec:experiments_category}{, i.e., $K=100$ 
for $\alpha_c$, $\rho_c$, and $\theta_u$, and $10$ latent features for
the seasonal and price vectors. We additionally}
tie the price vectors $\beta_c$ and seasonal effect vectors $\delta_c$
to all items in the same category.  To speed up computation, we fit
this model without thinking ahead.

We can again find similar items to ``query'' items using the cosine
distance between attribute vectors $\alpha_c$.
\Cref{tab:mostsimilar_upc} shows similar items for several queries;
the model identifies qualitatively related items.

\begin{figure}[t]
  \includegraphics[width=0.8\textwidth]{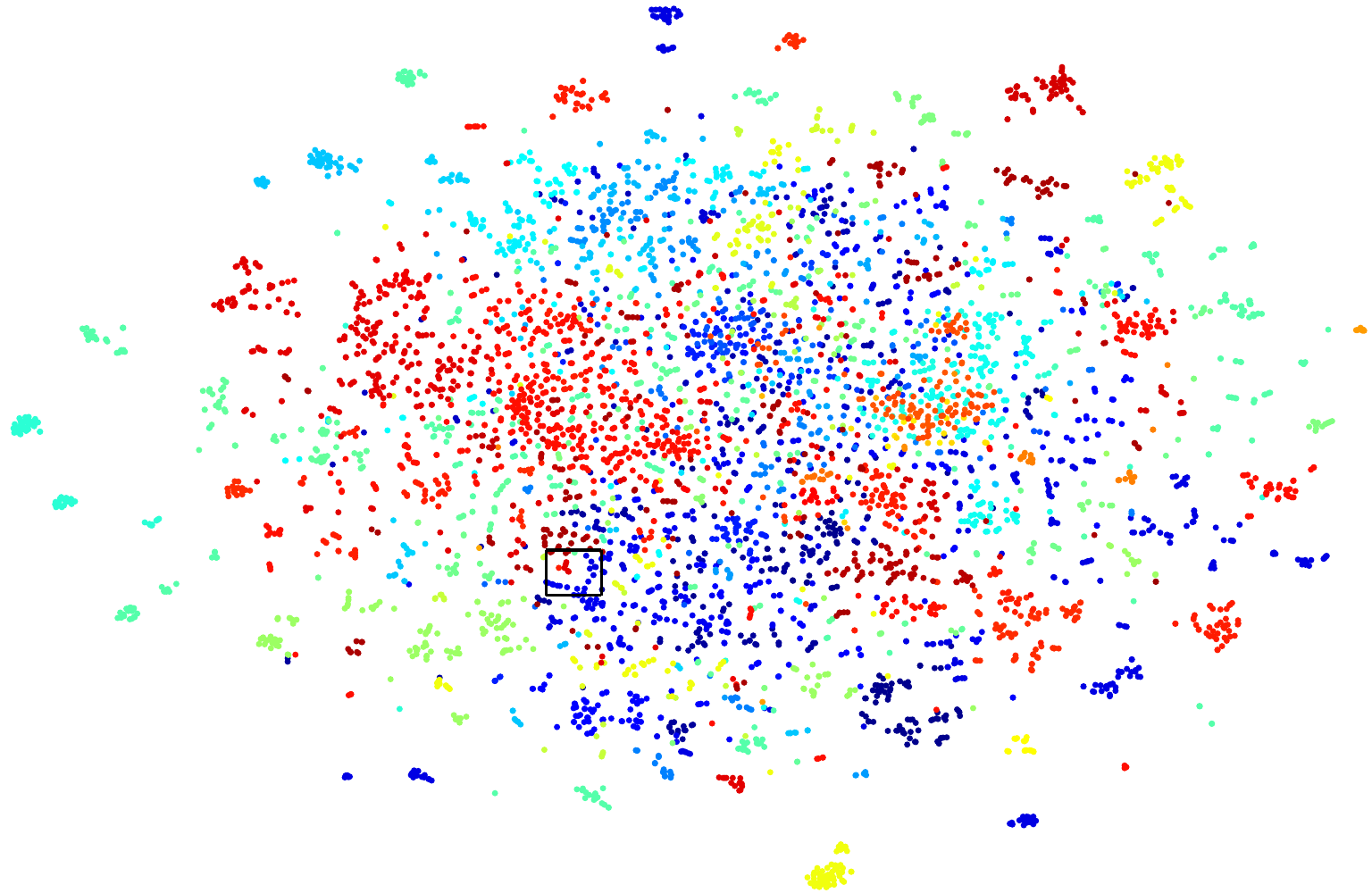}
  \caption{Two-dimensional \gls{tSNE} projection of the item feature vectors $\alpha_c$ for the \textsc{upc}-level experiment. Each item is represented with a dot whose color indicates its group (defined as one level of hierarchy above categories).\label{fig:tsne_upc1}}
\end{figure}

\begin{figure}[t]
  \includegraphics[width=1.0\textwidth]{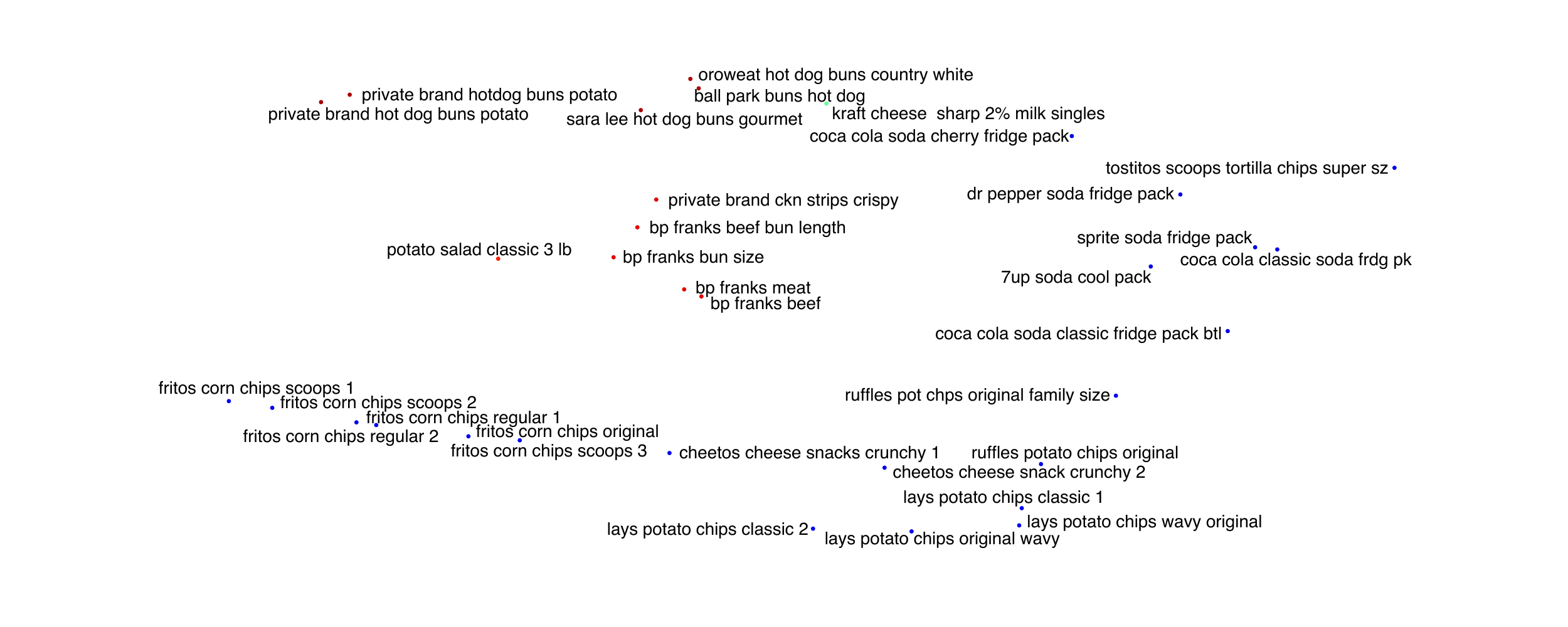}
  \caption{Zoom on the two-dimensional \gls{tSNE} projection of the item feature vectors, corresponding to the square indicated in \Cref{fig:tsne_upc1}.\label{fig:tsne_upc2}}
\end{figure}

For another view, \Cref{fig:tsne_upc1} shows a two-dimensional
\gls{tSNE} projection \citep{vanderMaaten2008} of the attribute
vectors. This figure colors the items according to their group,
and it reveals that items in the same category are often close to each
other in attribute space.
(Groups are
defined as one level of hierarchy above categories; some examples are
``jams, jellies, and spreads,'' ``salty snacks,'' or ``canned fruits.'')
When groups are mixed in a region, they
tend to be items that appear in similar
contexts, e.g., hot dogs, hamburger buns, and soda
(\Cref{fig:tsne_upc2}).

\begin{table}[t]
	\scriptsize
	\begin{tabular}{ccc} \toprule
		Dentyne ice gum peppermint			& california avocados				& Coca Cola classic soda fridge pack 1 \\ \midrule
		Wrigleys gum orbit white peppermint				& tomatoes red tov/cluster			& Sprite soda fridge pack \\
		Dentyne ice shivermint		& apples fuji medium 100ct	& Coca Cola classic fridge pack 2 \\
		Dentyne ice gum spearmint	& tomatoes roma red					& Coca Cola soda cherry fridge pack \\ \bottomrule
	\end{tabular}
	\caption{Most similar items to a query item (given in the first row), according to the cosine distance between the item features $\alpha_c$, for the \gls{UPC}-level experiment.}
	\label{tab:mostsimilar_upc}
\end{table}


\subsubsection{Substitutes, complements, and exchangeability metrics}
A key objective for applications of \shopper\ is to be able to
estimate interaction effects among products.  These effects are
described by the coefficients $\rho_c$ and attributes $\alpha_c$. When
$\rho_c^\top \alpha_{c'}$ and $\rho_{c'}^\top \alpha_c$ are large,
this means that purchasing item $c^\prime$ increases the consumer's
preference for $c$, and vice-versa.  When these two terms are negative
and large, the items may be substitutes---putting one item in the
basket reduces the need for the other item.

All else equal, complements are relatively likely to be purchased
together, while substitutes are less likely to be purchased
together. In addition, for complementary items $c$ and $c'$, when the
price of $c'$ increases, the customer is less likely to purchase item
$c$. We define the \emph{complementarity metric} as
\begin{align}
  C_{cc'} \triangleq \frac{1}{2}\left(\rho_c^\top \alpha_{c'} + \rho_{c'}^\top \alpha_{c} \right).
\end{align}


Most shopping data is sparse---each individual item has a low
probability of purchase---and so it can be difficult to accurately
identify pairs of items that are substitutes.  We introduce a new
metric, which we call \textit{exchangeability}, that can help identify
substitutes.  For a pair of items $(c,c')$, our notion of
exchangeability depends on the distributions of items that are induced
when conditioning on $c$ or $c'$; if those distributions are similar
then we say that $c$ and $c'$ are exchangeable.

Let $p_{k|c}$ denote the probability of item $k$ given that item $c$
is the only item in the basket (in this definition, we zero-out the probabilities
of items $c$, $c'$, and checkout). We measure the similarity between
the distributions $p_{\cdot|c}$ and $p_{\cdot|c'}$ with symmetrized
\gls{KL} divergence.  The \emph{exchangeability metric} is
\begin{align}
  E_{cc'} & \triangleq  \frac{1}{2}\left( D_{\textrm{KL}}\left(
            p_{\cdot|c} \;||\; p_{\cdot|c'} \right)  +
            D_{\textrm{KL}}\left( p_{\cdot|c'} \;||\; p_{\cdot|c}
            \right) \right) \\
          & =  \frac{1}{2} \sum_{k\neq c,c'} \left( p_{k|c}
            \log\frac{p_{k|c}}{p_{k|c'}} + p_{k|c'}
            \log\frac{p_{k|c'}}{p_{k|c}} \right).\nonumber
\end{align}
With this definition, two items that are intuitively
exchangeable---such as two different prepared sandwiches, or two
yogurts that are the same brand---will exhibit smaller values of
$E_{cc'}$.

With several example queries, \Cref{tab:CE_upc} shows the three most
complementary items and the three most exchangeable
items. In the computation of the probabilities $p_{k|c}$,
we consider an ``average customer'' (obtained by averaging the per-user vectors)
in an ``average week'' (obtained by averaging the per-week vectors). The
probabilities were also obtained assuming that no other item except $c$ is in
the shopping basket.
\shopper\ correctly recovers complements, such as taco shells and taco
seasoning, hot dogs and buns, and mustard and hamburger buns.
\shopper\ typically identifies the most exchangeable items as being of
the same type, such as several types of buns or tortillas; highly
exchangeable items are suggestive of substitutes.

{Finally, note that }\shopper{ can model triplets of complementary items
when they are also pairwise complements, but it cannot in cases where the triplets
do not form complementary pairs.}

\begin{table}[h]
  \scriptsize
  \begin{tabular}{crlrl} \toprule
    query items & \multicolumn{2}{l}{complementarity score} & \multicolumn{2}{l}{exchangeability score} \\ \midrule
    \multirow{3}{*}{\shortstack[c]{mission tortilla\\soft taco 1}}
                &	$2.40$ &	 taco bell taco seasoning mix 	& $0.05$ & mission fajita size \\
                &	$2.26$ &	 mcrmck seasoning mix taco 		& $0.07$ & mission tortilla soft taco 2 \\
                &	$2.24$ &	 lawrys taco seasoning mix 		& $0.13$ & mission tortilla fluffy gordita \\  \midrule
    \multirow{3}{*}{\shortstack[c]{private brand\\hot dog buns}}
                &	$2.99$ &	 bp franks meat 			& $0.11$ & ball park buns hot dog \\
                &	$2.63$ &	 bp franks bun size 		& $0.13$ & private brand hotdog buns potato 1 \\
                &	$2.37$ &	 bp franks beed bun length 			& $0.15$ & private brand hotdog buns potato 2 \\  \midrule
    \multirow{3}{*}{\shortstack[c]{private brand mustard\\squeeze bottle}}
                &	$0.50$ &	 private brand hot dog buns 			& $0.15$ & frenchs mustard classic yellow squeeze \\
                &	$0.41$ &	 private brand cutlery full size forks 	& $0.16$ & frenchs mustard classic yellow squeezed \\
                &	$0.24$ &	 best foods mayonnaise squeeze 				& $0.21$ & heinz ketchup squeeze bottle \\  \midrule
    \multirow{3}{*}{\shortstack[c]{private brand napkins\\all occasion}}
                &	$0.78$ &	 private brand selection plates 6 7/8 in 	& $0.09$ & vnty fair napkins all occasion 1 \\
                &	$0.50$ &	 private brand selection plates 8 3/4 in & $0.11$ & vnty fair napkins all occasion 2 \\
                &	$0.49$ &	 private brand cutlery full size forks 	& $0.12$ & private brand selection premium napkins \\  \bottomrule
  \end{tabular}
  \caption{Items with the highest complementarity and lowest exchangeability metrics for some query items.}
  \label{tab:CE_upc}
\end{table}



\section{Discussion}
\label{sec:conclusions}

We developed \shopper, a probabilistic model of consumer behavior.
\shopper\ is a sequential model of a consumer that posits a latent
structure to how he or she shops. In posterior inference, \shopper\
estimates item attributes, item-item interactions, price elasticities,
seasonal effects, and customer preferences.  We used \shopper\ to
analyze large-scale shopping data from a grocery store and evaluated
it with out-of-sample predictions.  In addition to evaluating on a
random sample of held out data, we also evaluated on counterfactual
test sets, i.e., on days where prices are systematically different
from average.


\shopper\ is a computationally feasible, interpretable approach to
modeling consumer choice over complex shopping baskets. While the
existing literature in economics and marketing usually considers two
or three products \citep{berry2014structural}, \shopper\ enables
considering choices among thousands of products that potentially
interact with one another.  Using supermarket shopping data, we show
that \shopper\ can uncover these relationships.


There are several avenues for future work.  Consider the question of
how the probabilistic model relates to maximizing a consumer's utility
of the entire basket.  \Cref{sec:basket_utility} introduced a
heuristic model of consumer behavior that is consistent with the
probabilistic model. This heuristic lends itself to computational
tractability; it enables us to analyze datasets of baskets that
involve thousands of products and thousands of consumers. But the
heuristic also involves a fairly myopic consumer. In future work, it
is interesting to consider alternative heuristic models.





Another avenue is to embellish the distribution of baskets.  One way
is to expand the model to capture within-basket
heterogeneity. Shopping trips may reflect a collection of needs, such
as school lunches, a dinner party, and pet care, and items may
interact with other items within each need.  Capturing the
heterogenous patterns within the baskets would sharpen the estimated
interactions.  {Another embellishment is to expand the model of
  baskets to include a budget.  A budget imposes constraints on the
  number of items (or their total price) purchased in a single trip.}




\section*{Acknowledgements}
Francisco J.\ R.\ Ruiz is supported by the EU H2020 programme (Marie Sk\l{}odowska-Curie Individual Fellowship, grant agreement 706760).
This work is also supported by ONR N00014-17-1-2131, NIH 1U01MH115727-01, DARPA SD2 FA8750-18-C-0130, ONR N00014-15-1-2209, NSF CCF-1740833, IBM, 2Sigma, and Amazon.
The authors also thank Tilman Drerup and Tobias Schmidt for research assistance,
the seminar participants at Stanford,
as well as the Cyber Initiative at Stanford.
Finally, we also acknowledge the support of NVIDIA Corporation with the donation of a GPU used for this research.


\newpage
\appendix

\section{Details on the Inference Algorithm}
\label{sec:supp_inference}

Here we provide the technical details of the variational inference procedure and a description of the full algorithm.

Recall the notation introduced in Section~5 of the main paper, where $\mb{\ell} = \{\rho,\alpha, \lambda, \theta, \gamma, \beta, \mu, \delta\}$ denotes the set of all latent variables in the model, $\mathpzc{y}=\{ \mathpzc{y}_t\}$ is the collection of (unordered) baskets, and $x=x_{1:T}$, where $x_t=(u_t,w_t,r_t)$ are observed covariates of the shopping trips. We use mean-field variational inference, i.e., we approximate the posterior $p(\mb{\ell} \g \mathpzc{y}, \bx)$ with a fully factorized variational distribution,
\begin{equation}
    \begin{split}
        q(\mb{\ell}) = & \prod_{c} q(\alpha_c)q(\rho_c)q(\lambda_c)q(\beta_c)q(\mu_c) \times \prod_{u} q(\theta_u)q(\gamma_u) \times \prod_{w} q(\delta_{w}).
    \end{split}
\end{equation}
We set each variational factor in the same family as the prior, i.e., Gaussian variational distributions with diagonal covariance matrices for $q(\rho_i)$, $q(\alpha_i)$, $q(\lambda_i)$, $q(\nu_u)$, $q(\mu_i)$, and $q(\delta_w)$, and independent gamma variational distributions for the price sensitivity terms $q(\gamma_u)$ and $q(\beta_i)$. We parameterize the Gaussian in terms of its mean and standard deviation, and we parameterize the gamma in terms of its shape and mean.

Let $\nu$ denote the vector containing all the variational parameters. We wish to find the variational parameters $\nu$ that maximize the \gls{ELBO},
\begin{equation}\label{eq-supp:elbo}
    \begin{split}
        \Lcal(\nu) & = \E{q(\mb{\ell} \prm \nu)}{\log p(\mathpzc{y} \g \bx,\mb{\ell}) + \log p(\mb{\ell}) - \log q(\mb{\ell} \prm \nu)} \\
        & = \mathbb{E}_{q(\mb{\ell} \prm \nu)}\left[ \sum_{t=1}^{T} \log p(\mathpzc{y}_t \g x_t,\mb{\ell}) + \log p(\mb{\ell}) - \log q(\mb{\ell} \prm \nu) \right] \\
        & \leq \log p(\mathpzc{y} \g \bx).
    \end{split}
\end{equation}

In this supplement, we show how to apply stochastic optimization to maximize the bound on the log marginal likelihood. More in detail, we first describe how to tackle the intractable expectations using stochastic optimization and the reparameterization trick, and then we show how to leverage stochastic optimization to decrease the computational complexity.

\subsection{Intractable expectations: Stochastic optimization and the reparameterization trick}
\label{sec:reparameterization}
We are interested in maximizing an objective function of the form
\begin{equation}\label{eq-supp:objective_reparam}
    \widetilde{\Lcal}(\nu) = \E{q(\mb{\ell} \prm \nu)}{f(\mb{\ell}, \nu)}
\end{equation}
with respect to the parameters $\nu$. In the particular case that $f(\mb{\ell},\nu)=\log p(\mathpzc{y},\mb{\ell} \g \bx)-\log q(\mb{\ell} \prm \nu)$, we recover the \gls{ELBO} in \Cref{eq-supp:elbo}, but we prefer to keep the notation general because in \Cref{sec:minibatches} we will consider other functions $f(\mb{\ell},\nu)$. 

The main challenge is that the expectations in \Cref{eq-supp:objective_reparam} are analytically intractable.
Thus, the variational algorithm we develop aims at obtaining and following noisy estimates of the gradient $\nabla_\nu \widetilde{\Lcal}$. We obtain these estimates via stochastic optimization; in particular, we apply the reparameterization trick to form Monte Carlo estimates of the gradient \citep{Kingma2014,Titsias2014_doubly,Rezende2014}.

In reparameterization, we first introduce a transformation of the latent variables $\mb{\ell}=\Tcal(\beps\prm\nu)$ and an auxiliary distribution $\pi(\beps\prm\nu)$, such that we can obtain samples from the variational distribution $q(\mb{\ell}\prm\nu)$ following a two-step process:
\begin{equation}\label{eq-supp:rep_transform}
    \beps\sim\pi(\beps\prm\nu),\qquad \mb{\ell} = \Tcal(\beps\prm\nu).
\end{equation}
The requirement for $\pi(\beps\prm\nu)$ and $\Tcal(\beps\prm\nu)$ is that this procedure must provide a variable $\mb{\ell}$ that is distributed according to $\mb{\ell}\sim q(\mb{\ell}\prm\nu)$. Here we have considered the generalized reparameterization approach \citep{Ruiz2016nips,Naesseth2017}, which allows the auxiliary distribution $\pi(\beps\prm\nu)$ to depend on the variational parameters $\nu$ (this is necessary because the gamma random variables are not otherwise reparameterizable).

Once we have introduced the auxiliary variable $\beps$, we can rewrite the gradient of the objective in \Cref{eq-supp:objective_reparam} as an expectation with respect to the auxiliary distribution $\pi(\beps\prm\nu)$,
\begin{equation}
    \nabla_{\nu}\widetilde{\Lcal} = \nabla_{\nu} \E{q(\mb{\ell}\prm\nu)}{f(\mb{\ell},\nu)}
    = \nabla_{\nu}\E{\pi(\beps\prm\nu)}{f\left(\Tcal(\beps\prm\nu),\nu\right)}.
\end{equation}
We now push the gradient into the integral\footnote{In the model that includes thinking-ahead, this step introduces a small bias due to the non-differentiability of the $\max(\cdot)$ operator; see \citet{Lee2018}.}
and apply the chain rule for derivatives to express the gradient as an expectation,
\begin{equation}\label{eq-supp:gradient_rep}
    \begin{split}
        \nabla_{\nu}\widetilde{\Lcal} = \E{\pi(\beps\prm\nu)}{\nabla_{\mb{\ell}} f(\mb{\ell},\nu)\big|_{\mb{\ell}=\Tcal(\beps\prm\nu)} \nabla_{\nu} \Tcal(\beps\prm\nu)
        + f\left(\Tcal(\beps\prm\nu),\nu\right)\nabla_{\nu} \log \pi(\beps\prm\nu) }.
    \end{split}
\end{equation}
To obtain this expression, we have assumed that $\E{\pi(\beps\prm\nu)}{\nabla_{\nu} f(\mb{\ell},\nu)\big|_{\mb{\ell}=\Tcal(\beps\prm\nu)}}=0$ because the only dependence of $f(\mb{\ell},\nu)$ on $\nu$ is through the term $\log q(\mb{\ell} \prm \nu)$, and the expectation of the score function is zero.

We can now obtain a Monte Carlo estimate of the expectation in \Cref{eq-supp:gradient_rep} (and therefore of the gradient of interest) by drawing a sample from $\pi(\beps\prm\nu)$ and evaluating the argument of the expectation. That is, we form the gradient estimator as
\begin{equation}\label{eq-supp:grad_estimate}
    \begin{split}
        \widehat{\nabla}_{\nu}\widetilde{\Lcal} = \nabla_{\mb{\ell}} f(\mb{\ell},\nu)\big|_{\mb{\ell}=\Tcal(\beps\prm\nu)} \nabla_{\nu} \Tcal(\beps\prm\nu) 
         + f\left(\Tcal(\beps\prm\nu),\nu\right)\nabla_{\nu} \log \pi(\beps\prm\nu),
    \end{split}
\end{equation}
where $\beps\sim\pi(\beps\prm\nu)$. This assumes that we are able to evaluate $f(\mb{\ell},\nu)$ and its gradient. (We show in \Cref{sec:minibatches} how to do that efficiently.)

We use a transformation $\Tcal(\cdot)$ and auxiliary distribution $\pi(\cdot)$ for each variational factor. For a Gaussian variational factor with mean $\mu$ and standard deviation $\sigma$, we use the standard reparameterization,
\begin{equation}
    \Tcal_{\textrm{Gauss}}(\varepsilon\prm\mu,\sigma)=\mu+\sigma\varepsilon, \qquad \pi_{\textrm{Gauss}}(\varepsilon\prm\mu,\sigma) = \Ncal(0,1),
\end{equation}
which makes the last term of \Cref{eq-supp:grad_estimate} vanish because $\nabla_{\nu}\log\pi_{\textrm{Gauss}}(\varepsilon\prm\mu,\sigma)=0$. For a gamma variational factor with shape $\alpha$ and mean $\mu$, we use the transformation based on rejection sampling \citep{Marsaglia2000},
\begin{equation}
    \Tcal_{\textrm{Gamma}}(\varepsilon\prm\alpha,\mu)= \frac{\mu}{\alpha}\left(\alpha-\frac{1}{3}\right)\left(1+\frac{\varepsilon}{\sqrt{9\alpha-3}}\right)^3,
\end{equation}
and $\pi_{\textrm{Gamma}}(\varepsilon\prm\alpha,\mu)$ is defined though a rejection sampling procedure. See \citet{Naesseth2017} for further details about the reparameterization trick for gamma random variables.\footnote{In particular, we also apply the ``shape augmentation trick,'' which allows us to reparameterize a gamma random variable with shape $\alpha$ in terms of another gamma random variable with shape $\alpha+P$, where $P$ is a positive integer. We use $P=10$. See \citet{Naesseth2017} for additional details.}

\Cref{alg:full_inference} summarizes the resulting variational inference procedure. At each iteration, we obtain a sample from $\mb{\ell}$ via the auxiliary distribution $\pi(\beps\prm\nu)$ and the transformation $\Tcal(\beps\prm\nu)$; we evaluate the function $f(\mb{\ell},\nu)$ and its gradient with respect to the latent variables $\mb{\ell}$; we obtain the gradient estimate in \Cref{eq-supp:grad_estimate}; and we take a gradient step for the variational parameters $\nu$. In the stochastic optimization procedure, we adaptively set the step size as proposed in the \gls{ADVI} algorithm \citep{Kucukelbir2017}.

\begin{algorithm}[t]
    \DontPrintSemicolon
    \SetAlgoLined
    \SetKwInOut{KwInput}{input}
    \SetKwInOut{KwOutput}{output}
    \KwInput{Data $\mathpzc{y}$ and $x$, model hyperparameters}
    \KwOutput{Variational parameters $\nu$}
    Initialize $\nu$ randomly\;
    Initialize iteration number $m\leftarrow 1$\;
    \Repeat{convergence} {
        Sample $\beps \sim \pi(\beps\prm\nu)$\;
        Compute $\mb{\ell} = \Tcal(\beps\prm\nu)$\;
        Evaluate $f(\mb{\ell},\nu)$ and $\nabla_{\mb{\ell}}f(\mb{\ell},\nu)$ (see \Cref{alg:model_eval})\;
        Obtain an estimate of the gradient, $\widehat{\nabla}_{\nu}\widetilde{\Lcal}$ (\Cref{eq-supp:grad_estimate})\;
        Set the step size $\eta^{(m)}$ (e.g., use the schedule proposed in \acrshort{ADVI})\;
        Take a gradient step, $\nu \leftarrow \nu + \eta^{(m)} \odot \widehat{\nabla}_{\nu}\widetilde{\Lcal}$\;
        Increase the iteration number, $m\leftarrow m+1$\;
    }
    \Return $\nu$\;
    \caption{Variational inference algorithm\label{alg:full_inference}}
\end{algorithm}

\subsection{Computational complexity: Stochastic optimization and lower bounds}
\label{sec:minibatches}
The algorithm in \Cref{sec:reparameterization} requires to evaluate the model log joint (as well as its gradient). There are three issues that make it expensive to evaluate the log joint. First, evaluating the log likelihood is expensive because it involves a summation over shopping trips. This represents a problem when the dataset is large. Second, evaluating the softmax involves computing its normalization constant, which contains a summation over all items. This becomes an issue when there are thousands of items and we need to evaluate many softmax probabilities. Third, computing the probability over unordered baskets $\mathpzc{y}_t$ is also expensive, as it involves a summation over all possible permutations.

We address these issues by combining two techniques: data subsampling and lower bounds on the \gls{ELBO}.
We first describe data subsampling for evaluating the log likelihood. Note that the log likelihood involves a summation over many terms,
\begin{equation}
    \log p(\mathpzc{y} \g \bx,\mb{\ell}) = \sum_{t=1}^{T} \log p(\mathpzc{y}_t \g x_t,\mb{\ell}).
\end{equation}
We can obtain an unbiased estimate of the log likelihood (and its gradient) by sampling a random subset of shopping trips. Let $\mathcal{B}_T$ be the (randomly chosen) set of trips. The estimator
\begin{equation*}
    \frac{T}{|\mathcal{B}_T|}\sum_{t\in \mathcal{B}_T} \log p(\mathpzc{y}_t \g x_t,\mb{\ell}) 
\end{equation*}
is unbiased because its expected value is the log likelihood $\log p(\mathpzc{y} \g \bx,\mb{\ell})$ \citep{Hoffman2013}. Thus, we subsample data terms to obtain unbiased estimates of the log likelihood and its gradient, resulting in a computationally more efficient algorithm.

Second, we describe how to form variational bounds to address the issue of the expensive normalization constant of the softmax. Each softmax log probability is given by
\begin{equation}
    \log p(y_{ti}=c\g \mb{y}_{t,i-1}) = \Psi(c, \mb{y}_{t,i-1})  - \log\left(\sum_{c^\prime \notin \mb{y}_{t,i-1}} \exp\{ \Psi(c^\prime, \mb{y}_{t,i-1}) \}\right).
\end{equation}
The summation over $c^\prime$ is expensive, and we cannot easily form an unbiased estimator because of the non-linearity introduced by the logarithm. Hence, we apply the one-vs-each bound \citep{Titsias2016}, which allows us to write
\begin{equation}
    \log p(y_{ti}=c\g \mb{y}_{t,i-1}) \geq \sum_{c^\prime \notin [\mb{y}_{t,i-1}, c] } \log \sigma \big(\Psi(c, \mb{y}_{t,i-1}) - \Psi(c^\prime, \mb{y}_{t,i-1})\big),
\end{equation}
where $\sigma(x)=\frac{1}{1+e^{-x}}$ is the sigmoid function. We can form unbiased estimates of the summation via subsampling. More precisely, we randomly sample a set $\mathcal{B}_{C}^{(t,i)}$ of items (each of them distinct from $c$ and from the other items in the basket). Then, we form the following unbiased estimator:
\begin{equation*}
    \frac{C-i}{|\mathcal{B}_{C}^{(t,i)}|} \sum_{c^\prime \in \mathcal{B}_{C}^{(t,i)}} \log \sigma \big(\Psi(c, \mb{y}_{t,i-1}) - \Psi(c^\prime, \mb{y}_{t,i-1})\big).
\end{equation*}
Here, $C$ stands for the total number of items.

Finally, we show how to deal with the issue of unordered baskets. Recall that each log likelihood term involves a summation over all possible permutations of the baskets (holding the checkout item in the last position),
\begin{equation}
    \log p(\mathpzc{y}_t \g x_t,\mb{\ell}) = \log \left( \sum_{\pi_t} p(\mb{y}_{t,\pi_t} \g x_t,\mb{\ell})\right).
\end{equation}
Following a similar procedure as \citet{DoshiVelez2009}, we introduce an auxiliary distribution $q(\pi_t)$ to rewrite the expression above as an expectation with respect to $q(\pi_t)$, and then we apply Jensen's inequality:
\begin{equation}
    \begin{split}
        \log p(\mathpzc{y}_t \g x_t,\mb{\ell}) & = \log \left( \E{q(\pi_t)}{\frac{p(\mb{y}_{t,\pi_t} \g x_t,\mb{\ell})}{q(\pi_t)}}\right) \\
        & \geq \E{q(\pi_t)}{\log p(\mb{y}_{t,\pi_t} \g x_t,\mb{\ell}) - \log q(\pi_t)} \\
        & = \sum_{\pi_t} q(\pi_t) \left( \log p(\mb{y}_{t,\pi_t} \g x_t,\mb{\ell}) - \log q(\pi_t) \right).
    \end{split}
\end{equation}
Since the bound involves a direct summation over permutations, we can subsample terms to alleviate the computational complexity. For simplicity, we set $q(\pi_t)$ to be a uniform distribution over all possible permutations, and thus we do not introduce auxiliary variational parameters that would be too expensive to obtain otherwise. In particular, we form an unbiased estimate of the bound by sampling one random permutation $\pi_t$ and evaluating the term\footnote{We ignore the term $\log q(\pi_t)$ because it is a constant.}
\begin{equation*}
        \log p(\mb{y}_{t,\pi_t} \g x_t,\mb{\ell}).
\end{equation*}

To sum up, we have derived a bound of the \gls{ELBO},
\begin{equation}
    \widetilde{\Lcal}(\nu)\leq \Lcal(\nu) \leq \log p(\mathpzc{y} \g x),
\end{equation}
which can still be written as an expectation with respect to the variational distribution $q(\mb{\ell}\g \nu)$. More importantly, we can efficiently evaluate the argument $f(\mb{\ell},\nu)$ of such expectation and its gradient with respect to the latent variables $\mb{\ell}$.

Putting all together, the function $f(\mb{\ell},\nu)$ that we use is given by
\begin{equation}\label{eq-supp:definition_f}
    \begin{split}
        f(\mb{\ell},\nu)= & \sum_{t=1}^{T} \sum_{\pi_t} q(\pi_t) \sum_{i=1}^{n_t}\sum_{c^\prime \notin [\mb{y}_{t,i-1}, c] } \log \sigma \big(\Psi(c, \mb{y}_{t,i-1}) - \Psi(c^\prime, \mb{y}_{t,i-1})\big) \\
        & + \log p(\mb{\ell})-\log q(\mb{\ell} \prm \nu).
    \end{split}
\end{equation}
We obtain an unbiased estimate via subsampling shopping trips, one permutation $\pi_t$ for each one, and items $c^\prime$. We use $|\mathcal{B}_T|=100$ trips and $|\mathcal{B}_{C}^{(t,i)}|=50$ items in our experiments.

\begin{algorithm}[t]
    \DontPrintSemicolon
    \SetAlgoLined
    \SetKwInOut{KwInput}{input}
    \SetKwInOut{KwOutput}{output}
    \KwInput{Data $\mathpzc{y}$ and $x$, a sample of the latent parameters $\mb{\ell}$, variational distribution $q(\mb{\ell}\prm\nu)$}
    \KwOutput{An unbiased estimate of $f(\mb{\ell},\nu)$ and its gradient (for the gradient, differentiate through the algorithm)}
    Initialize $\widehat{f} \leftarrow \log p(\mb{\ell}) - \log q(\mb{\ell}\prm\nu) $\;
    Sample a set of baskets $\mathcal{B}_T\subseteq\{1,\ldots,T\}$\;
    \For{$t\in \mathcal{B}_T$} {
        Sample a permutation $\pi_t$ of the items in basket $t$\;
        Set $\mb{y}_{t}$ to the vector containing the items in basket $t$ ordered according to $\pi_t$\;
        \For{$i=1,\ldots,n_t$} {
            Set $c$ to the $i$th item in $\mb{y}_{t}$ \;
            Sample a set of items $\mathcal{B}_{C}^{(t,i)}\subseteq\{1,\ldots,C\}\diagdown\{\mb{y}_{t,i-1}, c\}$\;
            \For{$c^\prime \in \mathcal{B}_{C}^{(t,i)}$} {
                Update $\widehat{f} \leftarrow \widehat{f} + \frac{T}{|\mathcal{B}_T|}\times\frac{C-i}{|\mathcal{B}_{C}^{(t,i)}|}\times \log\sigma\big(\Psi(c, \mb{y}_{t,i-1}) - \Psi(c^\prime, \mb{y}_{t,i-1})\big) $\;
            }
        }
    }
    \Return $\widehat{f}$ and $\nabla_{\mb{\ell}}\widehat{f}$\;
    \caption{Estimate of $f(\mb{\ell},\nu)$ (\Cref{eq-supp:definition_f})\label{alg:model_eval}}
\end{algorithm}

\Cref{alg:model_eval} outlines the procedure to obtain an unbiased estimate of $f(\mb{\ell},\nu)$ for a given sample of the latent parameters, $\mb{\ell}\sim q(\mb{\ell}\prm\nu)$. Differentiation though \Cref{alg:model_eval} gives the gradient $\nabla_{\mb{\ell}}f(\mb{\ell},\nu)$, which is also required in the inference procedure (\Cref{alg:full_inference}).


\bibliographystyle{imsart-nameyear}
\bibliography{fjrrLibrary}

\begin{thebibliography}{69}

\bibitem[\protect\citeauthoryear{Abernethy et~al.}{2009}]{Abernethy2009}
\begin{barticle}[author]
\bauthor{\bsnm{Abernethy},~\bfnm{J.}\binits{J.}},
  \bauthor{\bsnm{Bach},~\bfnm{F.}\binits{F.}},
  \bauthor{\bsnm{Evgeniou},~\bfnm{T.}\binits{T.}} \AND
  \bauthor{\bsnm{Vert},~\bfnm{J.~P.}\binits{J.~P.}}
(\byear{2009}).
\btitle{A New Approach to Collaborative Filtering: Operator Estimation with
  Spectral Regularization}.
\bjournal{Journal of Machine Learning Research}
\bvolume{10}
\bpages{803--826}.
\end{barticle}
\endbibitem

\bibitem[\protect\citeauthoryear{Arora et~al.}{2016}]{Arora2016}
\begin{barticle}[author]
\bauthor{\bsnm{Arora},~\bfnm{S.}\binits{S.}},
  \bauthor{\bsnm{Li},~\bfnm{Y.}\binits{Y.}},
  \bauthor{\bsnm{Liang},~\bfnm{Y.}\binits{Y.}} \AND
  \bauthor{\bsnm{Ma},~\bfnm{T.}\binits{T.}}
(\byear{2016}).
\btitle{{RAND-WALK: A} latent variable model approach to word embeddings}.
\bjournal{Transactions of the Association for Computational Linguistics}
\bvolume{4}.
\end{barticle}
\endbibitem

\bibitem[\protect\citeauthoryear{Athey and Stern}{1998}]{athey1998empirical}
\begin{btechreport}[author]
\bauthor{\bsnm{Athey},~\bfnm{Susan}\binits{S.}} \AND
  \bauthor{\bsnm{Stern},~\bfnm{Scott}\binits{S.}}
(\byear{1998}).
\btitle{An empirical framework for testing theories about complimentarity in
  organizational design}
\btype{Technical Report},
\bpublisher{National Bureau of Economic Research}.
\end{btechreport}
\endbibitem

\bibitem[\protect\citeauthoryear{Bamler and Mandt}{2017}]{Bamler2017}
\begin{binproceedings}[author]
\bauthor{\bsnm{Bamler},~\bfnm{R.}\binits{R.}} \AND
  \bauthor{\bsnm{Mandt},~\bfnm{S.}\binits{S.}}
(\byear{2017}).
\btitle{Dynamic word embeddings via skip-gram filtering}.
In \bbooktitle{International Conference in Machine Learning}.
\end{binproceedings}
\endbibitem

\bibitem[\protect\citeauthoryear{Barkan}{2016}]{Barkan2016bayesian}
\begin{barticle}[author]
\bauthor{\bsnm{Barkan},~\bfnm{O.}\binits{O.}}
(\byear{2016}).
\btitle{Bayesian neural word embedding}.
\bjournal{arXiv preprint arXiv:1603.06571}.
\end{barticle}
\endbibitem

\bibitem[\protect\citeauthoryear{Barkan and Koenigstein}{2016}]{Barkan2016}
\begin{binproceedings}[author]
\bauthor{\bsnm{Barkan},~\bfnm{O.}\binits{O.}} \AND
  \bauthor{\bsnm{Koenigstein},~\bfnm{N.}\binits{N.}}
(\byear{2016}).
\btitle{{Item2Vec}: Neural item embedding for collaborative filtering}.
In \bbooktitle{IEEE International Workshop on Machine Learning for Signal
  Processing}.
\end{binproceedings}
\endbibitem

\bibitem[\protect\citeauthoryear{Bengio et~al.}{2003}]{Bengio2003}
\begin{barticle}[author]
\bauthor{\bsnm{Bengio},~\bfnm{Y.}\binits{Y.}},
  \bauthor{\bsnm{Ducharme},~\bfnm{R.}\binits{R.}},
  \bauthor{\bsnm{Vincent},~\bfnm{P.}\binits{P.}} \AND
  \bauthor{\bsnm{Janvin},~\bfnm{C.}\binits{C.}}
(\byear{2003}).
\btitle{A Neural Probabilistic Language Model}.
\bjournal{Journal of Machine Learning Research}
\bvolume{3}
\bpages{1137--1155}.
\end{barticle}
\endbibitem

\bibitem[\protect\citeauthoryear{Bengio et~al.}{2006}]{bengio2006neural}
\begin{bincollection}[author]
\bauthor{\bsnm{Bengio},~\bfnm{Y.}\binits{Y.}},
  \bauthor{\bsnm{Schwenk},~\bfnm{H.}\binits{H.}},
  \bauthor{\bsnm{Sen\'{e}cal},~\bfnm{J.~S.}\binits{J.~S.}},
  \bauthor{\bsnm{Morin},~\bfnm{F.}\binits{F.}} \AND
  \bauthor{\bsnm{Gauvain},~\bfnm{J.~L.}\binits{J.~L.}}
(\byear{2006}).
\btitle{Neural probabilistic language models}.
In \bbooktitle{Innovations in Machine Learning}
\bpublisher{Springer}.
\end{bincollection}
\endbibitem

\bibitem[\protect\citeauthoryear{Berry et~al.}{2014}]{berry2014structural}
\begin{barticle}[author]
\bauthor{\bsnm{Berry},~\bfnm{Steven}\binits{S.}},
  \bauthor{\bsnm{Khwaja},~\bfnm{Ahmed}\binits{A.}},
  \bauthor{\bsnm{Kumar},~\bfnm{Vineet}\binits{V.}},
  \bauthor{\bsnm{Musalem},~\bfnm{Andres}\binits{A.}},
  \bauthor{\bsnm{Wilbur},~\bfnm{Kenneth~C}\binits{K.~C.}},
  \bauthor{\bsnm{Allenby},~\bfnm{Greg~M}\binits{G.~M.}},
  \bauthor{\bsnm{Anand},~\bfnm{Bharat~N}\binits{B.~N.}},
  \bauthor{\bsnm{Chintagunta},~\bfnm{Pradeep~K}\binits{P.~K.}},
  \bauthor{\bsnm{Hanemann},~\bfnm{W~Michael}\binits{W.~M.}},
  \bauthor{\bsnm{Jeziorski},~\bfnm{Przemyslaw}\binits{P.}} \betal{et~al.}
(\byear{2014}).
\btitle{Structural models of complementary choices}.
\bjournal{Marketing Letters}
\bvolume{25}
\bpages{245--256}.
\end{barticle}
\endbibitem

\bibitem[\protect\citeauthoryear{Blei, Kucukelbir and
  McAuliffe}{2017}]{Blei:2016}
\begin{barticle}[author]
\bauthor{\bsnm{Blei},~\bfnm{D.~M.}\binits{D.~M.}},
  \bauthor{\bsnm{Kucukelbir},~\bfnm{A.}\binits{A.}} \AND
  \bauthor{\bsnm{McAuliffe},~\bfnm{J.~D.}\binits{J.~D.}}
(\byear{2017}).
\btitle{Variational Inference: {A} Review for Statisticians}.
\bjournal{Journal of the American Statistical Association}
\bvolume{112}
\bpages{859--877}.
\end{barticle}
\endbibitem

\bibitem[\protect\citeauthoryear{Blum}{1954}]{Blum1954}
\begin{barticle}[author]
\bauthor{\bsnm{Blum},~\bfnm{J.~R.}\binits{J.~R.}}
(\byear{1954}).
\btitle{Approximation methods which converge with probability one}.
\bjournal{The Annals of Mathematical Statistics}
\bvolume{25}
\bpages{382--386}.
\end{barticle}
\endbibitem

\bibitem[\protect\citeauthoryear{Bottou, Curtis and
  Nocedal}{2016}]{Bottou:2016}
\begin{barticle}[author]
\bauthor{\bsnm{Bottou},~\bfnm{L{\'e}on}\binits{L.}},
  \bauthor{\bsnm{Curtis},~\bfnm{Frank~E.}\binits{F.~E.}} \AND
  \bauthor{\bsnm{Nocedal},~\bfnm{Jorge}\binits{J.}}
(\byear{2016}).
\btitle{Optimization Methods for Large-Scale Machine Learning}.
\bjournal{SIAM Review}
\bvolume{60}
\bpages{223--311}.
\end{barticle}
\endbibitem

\bibitem[\protect\citeauthoryear{Browning and
  Meghir}{1991}]{browning1991effects}
\begin{barticle}[author]
\bauthor{\bsnm{Browning},~\bfnm{Martin}\binits{M.}} \AND
  \bauthor{\bsnm{Meghir},~\bfnm{Costas}\binits{C.}}
(\byear{1991}).
\btitle{The effects of male and female labor supply on commodity demands}.
\bjournal{Econometrica: Journal of the Econometric Society}
\bpages{925--951}.
\end{barticle}
\endbibitem

\bibitem[\protect\citeauthoryear{Canny}{2004}]{Canny2004}
\begin{binproceedings}[author]
\bauthor{\bsnm{Canny},~\bfnm{J.}\binits{J.}}
(\byear{2004}).
\btitle{{GaP: A} factor model for discrete data}.
In \bbooktitle{Proceedings of the 27th Annual International ACM SIGIR
  Conference on Research and Development in Information Retrieval}.
\end{binproceedings}
\endbibitem

\bibitem[\protect\citeauthoryear{Cattell}{1952}]{Cattell1952}
\begin{bbook}[author]
\bauthor{\bsnm{Cattell},~\bfnm{R.~B.}\binits{R.~B.}}
(\byear{1952}).
\btitle{Factor Analysis: An Introduction and Manual for the Psychologist and
  Social Scientist}.
\bpublisher{New York: Harper}.
\end{bbook}
\endbibitem

\bibitem[\protect\citeauthoryear{Che, Chen and Chen}{2012}]{Che2012}
\begin{barticle}[author]
\bauthor{\bsnm{Che},~\bfnm{H.}\binits{H.}},
  \bauthor{\bsnm{Chen},~\bfnm{X.}\binits{X.}} \AND
  \bauthor{\bsnm{Chen},~\bfnm{Y.}\binits{Y.}}
(\byear{2012}).
\btitle{Investigating Effects of Out-of-Stock on Consumer Stockkeeping Unit
  Choice}.
\bjournal{Journal of Marketing Research}
\bvolume{49}
\bpages{502--513}.
\end{barticle}
\endbibitem

\bibitem[\protect\citeauthoryear{Chintagunta}{1994}]{chintagunta1994heterogeneous}
\begin{barticle}[author]
\bauthor{\bsnm{Chintagunta},~\bfnm{Pradeep~K}\binits{P.~K.}}
(\byear{1994}).
\btitle{Heterogeneous logit model implications for brand positioning}.
\bjournal{Journal of Marketing Research}
\bpages{304--311}.
\end{barticle}
\endbibitem

\bibitem[\protect\citeauthoryear{Chintagunta and
  Nair}{2011}]{chintagunta2011structural}
\begin{barticle}[author]
\bauthor{\bsnm{Chintagunta},~\bfnm{Pradeep~K}\binits{P.~K.}} \AND
  \bauthor{\bsnm{Nair},~\bfnm{Harikesh~S}\binits{H.~S.}}
(\byear{2011}).
\btitle{Structural workshop paper---Discrete-choice models of consumer demand
  in marketing}.
\bjournal{Marketing Science}
\bvolume{30}
\bpages{977--996}.
\end{barticle}
\endbibitem

\bibitem[\protect\citeauthoryear{Deaton and
  Muellbauer}{1980}]{deaton1980almost}
\begin{barticle}[author]
\bauthor{\bsnm{Deaton},~\bfnm{Angus}\binits{A.}} \AND
  \bauthor{\bsnm{Muellbauer},~\bfnm{John}\binits{J.}}
(\byear{1980}).
\btitle{An almost ideal demand system}.
\bjournal{The American economic review}
\bvolume{70}
\bpages{312--326}.
\end{barticle}
\endbibitem

\bibitem[\protect\citeauthoryear{Donnelly
  et~al.}{2019}]{athey2017counterfactual}
\begin{barticle}[author]
\bauthor{\bsnm{Donnelly},~\bfnm{R.}\binits{R.}},
  \bauthor{\bsnm{Ruiz},~\bfnm{F.~J.~R.}\binits{F.~J.~R.}},
  \bauthor{\bsnm{Blei},~\bfnm{D.~M.}\binits{D.~M.}} \AND
  \bauthor{\bsnm{Athey},~\bfnm{S.}\binits{S.}}
(\byear{2019}).
\btitle{Counterfactual Inference for Consumer Choice Across Many Product
  Categories}.
\bjournal{arXiv:1906.02635}.
\end{barticle}
\endbibitem

\bibitem[\protect\citeauthoryear{Doshi-Velez et~al.}{2009}]{DoshiVelez2009}
\begin{binproceedings}[author]
\bauthor{\bsnm{Doshi-Velez},~\bfnm{F.}\binits{F.}},
  \bauthor{\bsnm{Miller},~\bfnm{K.~T.}\binits{K.~T.}}, \bauthor{\bsnm{{Van
  Gael}},~\bfnm{J.}\binits{J.}} \AND
  \bauthor{\bsnm{Teh},~\bfnm{Y.~W.}\binits{Y.~W.}}
(\byear{2009}).
\btitle{Variational Inference for the {I}ndian Buffet Process}.
In \bbooktitle{Proceedings of the International Conference on Artificial
  Intelligence and Statistics}
\bvolume{12}.
\end{binproceedings}
\endbibitem

\bibitem[\protect\citeauthoryear{Elrod}{1988}]{elrod1988choice}
\begin{barticle}[author]
\bauthor{\bsnm{Elrod},~\bfnm{Terry}\binits{T.}}
(\byear{1988}).
\btitle{Choice map: Inferring a product-market map from panel data}.
\bjournal{Marketing Science}
\bvolume{7}
\bpages{21--40}.
\end{barticle}
\endbibitem

\bibitem[\protect\citeauthoryear{Elrod and Keane}{1995}]{elrod1995factor}
\begin{barticle}[author]
\bauthor{\bsnm{Elrod},~\bfnm{Terry}\binits{T.}} \AND
  \bauthor{\bsnm{Keane},~\bfnm{Michael~P}\binits{M.~P.}}
(\byear{1995}).
\btitle{A factor-analytic probit model for representing the market structure in
  panel data}.
\bjournal{Journal of Marketing Research}
\bpages{1--16}.
\end{barticle}
\endbibitem

\bibitem[\protect\citeauthoryear{Firth}{1957}]{Firth1957}
\begin{binproceedings}[author]
\bauthor{\bsnm{Firth},~\bfnm{J.~R.}\binits{J.~R.}}
(\byear{1957}).
\btitle{A synopsis of linguistic theory 1930-1955}.
In \bbooktitle{Studies in Linguistic Analysis (special volume of the
  Philological Society)}
\bvolume{1952--1959}.
\end{binproceedings}
\endbibitem

\bibitem[\protect\citeauthoryear{Gentzkow}{2007}]{gentzkow2007valuing}
\begin{barticle}[author]
\bauthor{\bsnm{Gentzkow},~\bfnm{Matthew}\binits{M.}}
(\byear{2007}).
\btitle{Valuing new goods in a model with complementarity: Online newspapers}.
\bjournal{The American Economic Review}
\bvolume{97}
\bpages{713--744}.
\end{barticle}
\endbibitem

\bibitem[\protect\citeauthoryear{Gopalan, Hofman and Blei}{2015}]{Gopalan2015}
\begin{binproceedings}[author]
\bauthor{\bsnm{Gopalan},~\bfnm{P.}\binits{P.}},
  \bauthor{\bsnm{Hofman},~\bfnm{J.}\binits{J.}} \AND
  \bauthor{\bsnm{Blei},~\bfnm{D.~M.}\binits{D.~M.}}
(\byear{2015}).
\btitle{Scalable recommendation with hierarchical {P}oisson factorization}.
In \bbooktitle{Uncertainty in Artificial Intelligence}.
\end{binproceedings}
\endbibitem

\bibitem[\protect\citeauthoryear{Gopalan et~al.}{2014}]{Gopalan2014}
\begin{binproceedings}[author]
\bauthor{\bsnm{Gopalan},~\bfnm{P.}\binits{P.}},
  \bauthor{\bsnm{Ruiz},~\bfnm{F.~J.~R.}\binits{F.~J.~R.}},
  \bauthor{\bsnm{Ranganath},~\bfnm{R.}\binits{R.}} \AND
  \bauthor{\bsnm{Blei},~\bfnm{D.~M.}\binits{D.~M.}}
(\byear{2014}).
\btitle{Bayesian Nonparametric {P}oisson Factorization for Recommendation
  Systems}.
In \bbooktitle{Artificial Intelligence and Statistics}.
\end{binproceedings}
\endbibitem

\bibitem[\protect\citeauthoryear{{G\"or\"ur}, J\"{a}kel and
  Rasmussen}{2006}]{Gorur2006}
\begin{binproceedings}[author]
\bauthor{\bsnm{{G\"or\"ur}},~\bfnm{D.}\binits{D.}},
  \bauthor{\bsnm{J\"{a}kel},~\bfnm{F.}\binits{F.}} \AND
  \bauthor{\bsnm{Rasmussen},~\bfnm{C.~E.}\binits{C.~E.}}
(\byear{2006}).
\btitle{A Choice Model with Infinitely Many Latent Features}.
In \bbooktitle{International Conference on Machine Learning}.
\end{binproceedings}
\endbibitem

\bibitem[\protect\citeauthoryear{Harris}{1954}]{Harris1954}
\begin{barticle}[author]
\bauthor{\bsnm{Harris},~\bfnm{Z.~S.}\binits{Z.~S.}}
(\byear{1954}).
\btitle{Distributional structure}.
\bjournal{Word}
\bvolume{10}
\bpages{146--162}.
\end{barticle}
\endbibitem

\bibitem[\protect\citeauthoryear{Hoffman et~al.}{2013}]{Hoffman2013}
\begin{barticle}[author]
\bauthor{\bsnm{Hoffman},~\bfnm{M.~D.}\binits{M.~D.}},
  \bauthor{\bsnm{Blei},~\bfnm{D.~M.}\binits{D.~M.}},
  \bauthor{\bsnm{Wang},~\bfnm{C.}\binits{C.}} \AND
  \bauthor{\bsnm{Paisley},~\bfnm{J.}\binits{J.}}
(\byear{2013}).
\btitle{Stochastic Variational Inference}.
\bjournal{Journal of Machine Learning Research}
\bvolume{14}
\bpages{1303--1347}.
\end{barticle}
\endbibitem

\bibitem[\protect\citeauthoryear{Hotz and Miller}{1993}]{hotz1993conditional}
\begin{barticle}[author]
\bauthor{\bsnm{Hotz},~\bfnm{V~Joseph}\binits{V.~J.}} \AND
  \bauthor{\bsnm{Miller},~\bfnm{Robert~A}\binits{R.~A.}}
(\byear{1993}).
\btitle{Conditional choice probabilities and the estimation of dynamic models}.
\bjournal{The Review of Economic Studies}
\bvolume{60}
\bpages{497--529}.
\end{barticle}
\endbibitem

\bibitem[\protect\citeauthoryear{Hu, Koren and Volinsky}{2008}]{Hu2008}
\begin{binproceedings}[author]
\bauthor{\bsnm{Hu},~\bfnm{Y.}\binits{Y.}},
  \bauthor{\bsnm{Koren},~\bfnm{Y.}\binits{Y.}} \AND
  \bauthor{\bsnm{Volinsky},~\bfnm{C.}\binits{C.}}
(\byear{2008}).
\btitle{Collaborative filtering for implicit feedback datasets}.
In \bbooktitle{IEEE International Conference on Data Mining}.
\end{binproceedings}
\endbibitem

\bibitem[\protect\citeauthoryear{Jordan et~al.}{1999}]{Jordan1999}
\begin{barticle}[author]
\bauthor{\bsnm{Jordan},~\bfnm{M.~I.}\binits{M.~I.}},
  \bauthor{\bsnm{Ghahramani},~\bfnm{Z.}\binits{Z.}},
  \bauthor{\bsnm{Jaakkola},~\bfnm{T.~S.}\binits{T.~S.}} \AND
  \bauthor{\bsnm{Saul},~\bfnm{L.~K.}\binits{L.~K.}}
(\byear{1999}).
\btitle{An Introduction to Variational Methods for Graphical Models}.
\bjournal{Machine Learning}
\bvolume{37}
\bpages{183--233}.
\end{barticle}
\endbibitem

\bibitem[\protect\citeauthoryear{Keane et~al.}{2013}]{keane2013panel}
\begin{barticle}[author]
\bauthor{\bsnm{Keane},~\bfnm{Michael~P}\binits{M.~P.}} \betal{et~al.}
(\byear{2013}).
\btitle{Panel data discrete choice models of consumer demand}.
\bjournal{Prepared for The Oxford Handbooks: Panel Data}.
\end{barticle}
\endbibitem

\bibitem[\protect\citeauthoryear{Kingma and Welling}{2014}]{Kingma2014}
\begin{binproceedings}[author]
\bauthor{\bsnm{Kingma},~\bfnm{D.~P.}\binits{D.~P.}} \AND
  \bauthor{\bsnm{Welling},~\bfnm{M.}\binits{M.}}
(\byear{2014}).
\btitle{Auto-Encoding Variational {B}ayes}.
In \bbooktitle{International Conference on Learning Representations}.
\end{binproceedings}
\endbibitem

\bibitem[\protect\citeauthoryear{Kucukelbir et~al.}{2017}]{Kucukelbir2017}
\begin{barticle}[author]
\bauthor{\bsnm{Kucukelbir},~\bfnm{A.}\binits{A.}},
  \bauthor{\bsnm{Tran},~\bfnm{D.}\binits{D.}},
  \bauthor{\bsnm{Ranganath},~\bfnm{R.}\binits{R.}},
  \bauthor{\bsnm{Gelman},~\bfnm{A.}\binits{A.}} \AND
  \bauthor{\bsnm{Blei},~\bfnm{D.~M.}\binits{D.~M.}}
(\byear{2017}).
\btitle{Automatic Differentiation Variational Inference}.
\bjournal{Journal of Machine Learning Research}
\bvolume{18}
\bpages{1--45}.
\end{barticle}
\endbibitem

\bibitem[\protect\citeauthoryear{Lee, Yu and Yang}{2018}]{Lee2018}
\begin{binproceedings}[author]
\bauthor{\bsnm{Lee},~\bfnm{W.}\binits{W.}},
  \bauthor{\bsnm{Yu},~\bfnm{H.}\binits{H.}} \AND
  \bauthor{\bsnm{Yang},~\bfnm{H.}\binits{H.}}
(\byear{2018}).
\btitle{Reparameterization Gradient for Non-differentiable Models}.
In \bbooktitle{Advances in Neural Information Processing Systems}.
\end{binproceedings}
\endbibitem

\bibitem[\protect\citeauthoryear{Levy and Goldberg}{2014}]{Levy2014neural}
\begin{binproceedings}[author]
\bauthor{\bsnm{Levy},~\bfnm{O.}\binits{O.}} \AND
  \bauthor{\bsnm{Goldberg},~\bfnm{Y.}\binits{Y.}}
(\byear{2014}).
\btitle{Neural word embedding as implicit matrix factorization}.
In \bbooktitle{Advances in Neural Information Processing Systems}.
\end{binproceedings}
\endbibitem

\bibitem[\protect\citeauthoryear{Liang et~al.}{2016}]{Liang2016}
\begin{binproceedings}[author]
\bauthor{\bsnm{Liang},~\bfnm{D.}\binits{D.}},
  \bauthor{\bsnm{Altosaar},~\bfnm{J.}\binits{J.}},
  \bauthor{\bsnm{Charlin},~\bfnm{L.}\binits{L.}} \AND
  \bauthor{\bsnm{Blei},~\bfnm{D.~M.}\binits{D.~M.}}
(\byear{2016}).
\btitle{Factorization Meets the Item Embedding: Regularizing Matrix
  Factorization with Item Co-occurrence}.
In \bbooktitle{ACM Conference on Recommender System}.
\end{binproceedings}
\endbibitem

\bibitem[\protect\citeauthoryear{Ma et~al.}{2011}]{Ma2011}
\begin{binproceedings}[author]
\bauthor{\bsnm{Ma},~\bfnm{H.}\binits{H.}},
  \bauthor{\bsnm{Liu},~\bfnm{C.}\binits{C.}},
  \bauthor{\bsnm{King},~\bfnm{I.}\binits{I.}} \AND
  \bauthor{\bsnm{R.},~\bfnm{Lyu~M.}\binits{L.~M.}}
(\byear{2011}).
\btitle{Probabilistic factor models for web site recommendation}.
In \bbooktitle{Proceedings of the 34th International ACM SIGIR Conference on
  Research and Development in Information Retrieval}.
\end{binproceedings}
\endbibitem

\bibitem[\protect\citeauthoryear{Marsaglia and Tang}{2000}]{Marsaglia2000}
\begin{barticle}[author]
\bauthor{\bsnm{Marsaglia},~\bfnm{G.}\binits{G.}} \AND
  \bauthor{\bsnm{Tang},~\bfnm{W.~W.}\binits{W.~W.}}
(\byear{2000}).
\btitle{A simple method for generating gamma variables}.
\bjournal{ACM Transactions on Mathematical Software}
\bvolume{26}
\bpages{363--372}.
\end{barticle}
\endbibitem

\bibitem[\protect\citeauthoryear{Mikolov, Yih and
  Zweig}{2013}]{Mikolov2013linguistic}
\begin{binproceedings}[author]
\bauthor{\bsnm{Mikolov},~\bfnm{T.}\binits{T.}},
  \bauthor{\bsnm{Yih},~\bfnm{W.~T.~au}\binits{W.~T.~a.}} \AND
  \bauthor{\bsnm{Zweig},~\bfnm{G.}\binits{G.}}
(\byear{2013}).
\btitle{Linguistic Regularities in Continuous Space Word Representations}.
In \bbooktitle{Conference of the North American Chapter of the Association for
  Computational Linguistics: Human Language Technologies}.
\end{binproceedings}
\endbibitem

\bibitem[\protect\citeauthoryear{Mikolov
  et~al.}{2013a}]{Mikolov2013distributed}
\begin{binproceedings}[author]
\bauthor{\bsnm{Mikolov},~\bfnm{T.}\binits{T.}},
  \bauthor{\bsnm{Sutskever},~\bfnm{I.}\binits{I.}},
  \bauthor{\bsnm{Chen},~\bfnm{K.}\binits{K.}},
  \bauthor{\bsnm{Corrado},~\bfnm{G.~S.}\binits{G.~S.}} \AND
  \bauthor{\bsnm{Dean},~\bfnm{J.}\binits{J.}}
(\byear{2013}a).
\btitle{Distributed representations of words and phrases and their
  compositionality}.
In \bbooktitle{Advances in Neural Information Processing Systems}.
\end{binproceedings}
\endbibitem

\bibitem[\protect\citeauthoryear{Mikolov et~al.}{2013b}]{Mikolov2013efficient}
\begin{barticle}[author]
\bauthor{\bsnm{Mikolov},~\bfnm{T.}\binits{T.}},
  \bauthor{\bsnm{Chen},~\bfnm{K.}\binits{K.}},
  \bauthor{\bsnm{Corrado},~\bfnm{G.~S.}\binits{G.~S.}} \AND
  \bauthor{\bsnm{Dean},~\bfnm{J.}\binits{J.}}
(\byear{2013}b).
\btitle{Efficient estimation of word representations in vector space}.
\bjournal{International Conference on Learning Representations}.
\end{barticle}
\endbibitem

\bibitem[\protect\citeauthoryear{Mnih and Hinton}{2007}]{Mnih2007}
\begin{binproceedings}[author]
\bauthor{\bsnm{Mnih},~\bfnm{A.}\binits{A.}} \AND
  \bauthor{\bsnm{Hinton},~\bfnm{G.~E.}\binits{G.~E.}}
(\byear{2007}).
\btitle{Three New Graphical Models for Statistical Language Modelling}.
In \bbooktitle{International Conference on Machine Learning}.
\end{binproceedings}
\endbibitem

\bibitem[\protect\citeauthoryear{Mnih and Kavukcuoglu}{2013}]{Mnih2013learning}
\begin{binproceedings}[author]
\bauthor{\bsnm{Mnih},~\bfnm{A.}\binits{A.}} \AND
  \bauthor{\bsnm{Kavukcuoglu},~\bfnm{K.}\binits{K.}}
(\byear{2013}).
\btitle{Learning word embeddings efficiently with noise-contrastive
  estimation}.
In \bbooktitle{Advances in Neural Information Processing Systems}.
\end{binproceedings}
\endbibitem

\bibitem[\protect\citeauthoryear{Mnih and Teh}{2012}]{Mnih2012}
\begin{binproceedings}[author]
\bauthor{\bsnm{Mnih},~\bfnm{A.}\binits{A.}} \AND
  \bauthor{\bsnm{Teh},~\bfnm{Y.~W.}\binits{Y.~W.}}
(\byear{2012}).
\btitle{A fast and simple algorithm for training neural probabilistic language
  models}.
In \bbooktitle{International Conference on Machine Learning}.
\end{binproceedings}
\endbibitem

\bibitem[\protect\citeauthoryear{Naesseth et~al.}{2017}]{Naesseth2017}
\begin{binproceedings}[author]
\bauthor{\bsnm{Naesseth},~\bfnm{C.}\binits{C.}},
  \bauthor{\bsnm{Ruiz},~\bfnm{F.~J.~R.}\binits{F.~J.~R.}},
  \bauthor{\bsnm{Linderman},~\bfnm{S.}\binits{S.}} \AND
  \bauthor{\bsnm{Blei},~\bfnm{D.~M.}\binits{D.~M.}}
(\byear{2017}).
\btitle{Reparameterization gradients through acceptance-rejection methods}.
In \bbooktitle{Artificial Intelligence and Statistics}.
\end{binproceedings}
\endbibitem

\bibitem[\protect\citeauthoryear{Ng and Russell}{2000}]{Ng2000}
\begin{binproceedings}[author]
\bauthor{\bsnm{Ng},~\bfnm{A.~Y.}\binits{A.~Y.}} \AND
  \bauthor{\bsnm{Russell},~\bfnm{S.~J.}\binits{S.~J.}}
(\byear{2000}).
\btitle{Algorithms for inverse reinforcement learning}.
In \bbooktitle{International Conference in Machine Learning}.
\end{binproceedings}
\endbibitem

\bibitem[\protect\citeauthoryear{Pennington, Socher and
  Manning}{2014}]{Pennington2014}
\begin{binproceedings}[author]
\bauthor{\bsnm{Pennington},~\bfnm{J.}\binits{J.}},
  \bauthor{\bsnm{Socher},~\bfnm{R.}\binits{R.}} \AND
  \bauthor{\bsnm{Manning},~\bfnm{C.~D.}\binits{C.~D.}}
(\byear{2014}).
\btitle{{GloVe: G}lobal Vectors for Word Representation}.
In \bbooktitle{Conference on Empirical Methods on Natural Language Processing}.
\end{binproceedings}
\endbibitem

\bibitem[\protect\citeauthoryear{Rezende, Mohamed and
  Wierstra}{2014}]{Rezende2014}
\begin{binproceedings}[author]
\bauthor{\bsnm{Rezende},~\bfnm{D.~J.}\binits{D.~J.}},
  \bauthor{\bsnm{Mohamed},~\bfnm{S.}\binits{S.}} \AND
  \bauthor{\bsnm{Wierstra},~\bfnm{D.}\binits{D.}}
(\byear{2014}).
\btitle{Stochastic backpropagation and approximate inference in deep generative
  models}.
In \bbooktitle{International Conference on Machine Learning}.
\end{binproceedings}
\endbibitem

\bibitem[\protect\citeauthoryear{Robbins and Monro}{1951}]{Robbins1951}
\begin{barticle}[author]
\bauthor{\bsnm{Robbins},~\bfnm{H.}\binits{H.}} \AND
  \bauthor{\bsnm{Monro},~\bfnm{S.}\binits{S.}}
(\byear{1951}).
\btitle{A stochastic approximation method}.
\bjournal{The Annals of Mathematical Statistics}
\bvolume{22}
\bpages{400--407}.
\end{barticle}
\endbibitem

\bibitem[\protect\citeauthoryear{Rudolph et~al.}{2016}]{Rudolph2016}
\begin{binproceedings}[author]
\bauthor{\bsnm{Rudolph},~\bfnm{M.}\binits{M.}},
  \bauthor{\bsnm{Ruiz},~\bfnm{F.~J.~R.}\binits{F.~J.~R.}},
  \bauthor{\bsnm{Mandt},~\bfnm{S.}\binits{S.}} \AND
  \bauthor{\bsnm{Blei},~\bfnm{D.~M.}\binits{D.~M.}}
(\byear{2016}).
\btitle{Exponential Family Embeddings}.
In \bbooktitle{Advances in Neural Information Processing Systems}.
\end{binproceedings}
\endbibitem

\bibitem[\protect\citeauthoryear{Ruiz, Athey and Blei}{2017}]{Ruiz2017_supp}
\begin{bmisc}[author]
\bauthor{\bsnm{Ruiz},~\bfnm{F.~J.~R.}\binits{F.~J.~R.}},
  \bauthor{\bsnm{Athey},~\bfnm{S.}\binits{S.}} \AND
  \bauthor{\bsnm{Blei},~\bfnm{D.~M.}\binits{D.~M.}}
(\byear{2017}).
\btitle{Supplement to ``{SHOPPER:} A Probabilistic Model of Consumer Choice
  with Substitutes and Complements''}.
\bhowpublished{DOI: 10.1214/00-AOASXXXXSUPP}.
\end{bmisc}
\endbibitem

\bibitem[\protect\citeauthoryear{Ruiz, Titsias and Blei}{2016}]{Ruiz2016nips}
\begin{binproceedings}[author]
\bauthor{\bsnm{Ruiz},~\bfnm{F.~J.~R.}\binits{F.~J.~R.}},
  \bauthor{\bsnm{Titsias},~\bfnm{M.~K.}\binits{M.~K.}} \AND
  \bauthor{\bsnm{Blei},~\bfnm{D.~M.}\binits{D.~M.}}
(\byear{2016}).
\btitle{The Generalized Reparameterization Gradient}.
In \bbooktitle{Advances in Neural Information Processing Systems}.
\end{binproceedings}
\endbibitem

\bibitem[\protect\citeauthoryear{Ruiz et~al.}{2018}]{Ruiz2018}
\begin{binproceedings}[author]
\bauthor{\bsnm{Ruiz},~\bfnm{F.~J.~R.}\binits{F.~J.~R.}},
  \bauthor{\bsnm{Titsias},~\bfnm{M.~K.}\binits{M.~K.}},
  \bauthor{\bsnm{Dieng},~\bfnm{A.~B.}\binits{A.~B.}} \AND
  \bauthor{\bsnm{Blei},~\bfnm{D.~M.}\binits{D.~M.}}
(\byear{2018}).
\btitle{Augment and Reduce: Stochastic Inference for Large Categorical
  Distributions}.
In \bbooktitle{International Conference on Machine Learning}.
\end{binproceedings}
\endbibitem

\bibitem[\protect\citeauthoryear{Russell}{1998}]{Russell1998}
\begin{binproceedings}[author]
\bauthor{\bsnm{Russell},~\bfnm{S.~J.}\binits{S.~J.}}
(\byear{1998}).
\btitle{Learning agents for uncertain environments}.
In \bbooktitle{Annual Conference on Computational Learning Theory}.
\end{binproceedings}
\endbibitem

\bibitem[\protect\citeauthoryear{Semenova et~al.}{2018}]{Semenova2018}
\begin{barticle}[author]
\bauthor{\bsnm{Semenova},~\bfnm{V.}\binits{V.}},
  \bauthor{\bsnm{Goldman},~\bfnm{M.}\binits{M.}},
  \bauthor{\bsnm{Chernozhukov},~\bfnm{V.}\binits{V.}} \AND
  \bauthor{\bsnm{Taddy},~\bfnm{M.}\binits{M.}}
(\byear{2018}).
\btitle{Orthogonal {ML} for Demand Estimation: High Dimensional Causal
  Inference in Dynamic Panels}.
\bjournal{arXiv:1712.09988}.
\end{barticle}
\endbibitem

\bibitem[\protect\citeauthoryear{Song and Chintagunta}{2007}]{song2007discrete}
\begin{barticle}[author]
\bauthor{\bsnm{Song},~\bfnm{Inseong}\binits{I.}} \AND
  \bauthor{\bsnm{Chintagunta},~\bfnm{Pradeep~K}\binits{P.~K.}}
(\byear{2007}).
\btitle{A discrete--continuous model for multicategory purchase behavior of
  households}.
\bjournal{Journal of Marketing Research}
\bvolume{44}
\bpages{595--612}.
\end{barticle}
\endbibitem

\bibitem[\protect\citeauthoryear{Stern, Herbrich and Thore}{2009}]{Stern2009}
\begin{binproceedings}[author]
\bauthor{\bsnm{Stern},~\bfnm{D.~H.}\binits{D.~H.}},
  \bauthor{\bsnm{Herbrich},~\bfnm{R.}\binits{R.}} \AND
  \bauthor{\bsnm{Thore},~\bfnm{G.}\binits{G.}}
(\byear{2009}).
\btitle{Matchbox: Large Scale Bayesian Recommendations}.
In \bbooktitle{18th International World Wide Web Conference}.
\end{binproceedings}
\endbibitem

\bibitem[\protect\citeauthoryear{Titsias}{2016}]{Titsias2016}
\begin{binproceedings}[author]
\bauthor{\bsnm{Titsias},~\bfnm{M.~K.}\binits{M.~K.}}
(\byear{2016}).
\btitle{One-vs-Each Approximation to Softmax for Scalable Estimation of
  Probabilities}.
In \bbooktitle{Advances in Neural Information Processing Systems}.
\end{binproceedings}
\endbibitem

\bibitem[\protect\citeauthoryear{Titsias and
  L\'{a}zaro-Gredilla}{2014}]{Titsias2014_doubly}
\begin{binproceedings}[author]
\bauthor{\bsnm{Titsias},~\bfnm{M.~K.}\binits{M.~K.}} \AND
  \bauthor{\bsnm{L\'{a}zaro-Gredilla},~\bfnm{M.}\binits{M.}}
(\byear{2014}).
\btitle{Doubly stochastic variational {B}ayes for non-conjugate inference}.
In \bbooktitle{International Conference on Machine Learning}.
\end{binproceedings}
\endbibitem

\bibitem[\protect\citeauthoryear{Train, McFadden and
  Ben-Akiva}{1987}]{train1987demand}
\begin{barticle}[author]
\bauthor{\bsnm{Train},~\bfnm{Kenneth~E}\binits{K.~E.}},
  \bauthor{\bsnm{McFadden},~\bfnm{Daniel~L}\binits{D.~L.}} \AND
  \bauthor{\bsnm{Ben-Akiva},~\bfnm{Moshe}\binits{M.}}
(\byear{1987}).
\btitle{The demand for local telephone service: A fully discrete model of
  residential calling patterns and service choices}.
\bjournal{The RAND Journal of Economics}
\bpages{109--123}.
\end{barticle}
\endbibitem

\bibitem[\protect\citeauthoryear{van~der Maaten and
  Hinton}{2008}]{vanderMaaten2008}
\begin{barticle}[author]
\bauthor{\bparticle{van~der} \bsnm{Maaten},~\bfnm{L.~J.~P.}\binits{L.~J.~P.}}
  \AND \bauthor{\bsnm{Hinton},~\bfnm{G.~E.}\binits{G.~E.}}
(\byear{2008}).
\btitle{Visualizing high-dimensional data using {t-SNE}}.
\bjournal{Journal of Machine Learning Research}
\bvolume{9}
\bpages{2579--2605}.
\end{barticle}
\endbibitem

\bibitem[\protect\citeauthoryear{Vilnis and McCallum}{2015}]{Vilnis2014}
\begin{binproceedings}[author]
\bauthor{\bsnm{Vilnis},~\bfnm{L.}\binits{L.}} \AND
  \bauthor{\bsnm{McCallum},~\bfnm{A.}\binits{A.}}
(\byear{2015}).
\btitle{Word representations via {G}aussian embedding}.
In \bbooktitle{International Conference on Learning Representations}.
\end{binproceedings}
\endbibitem

\bibitem[\protect\citeauthoryear{Wainwright and Jordan}{2008}]{Wainwright2008}
\begin{barticle}[author]
\bauthor{\bsnm{Wainwright},~\bfnm{M.~J.}\binits{M.~J.}} \AND
  \bauthor{\bsnm{Jordan},~\bfnm{M.~I.}\binits{M.~I.}}
(\byear{2008}).
\btitle{Graphical Models, Exponential Families, and Variational Inference}.
\bjournal{Foundations and Trends in Machine Learning}
\bvolume{1}
\bpages{1--305}.
\end{barticle}
\endbibitem

\bibitem[\protect\citeauthoryear{Wan et~al.}{2017}]{Wan2017}
\begin{binproceedings}[author]
\bauthor{\bsnm{Wan},~\bfnm{M.}\binits{M.}},
  \bauthor{\bsnm{Wang},~\bfnm{D.}\binits{D.}},
  \bauthor{\bsnm{Goldman},~\bfnm{M.}\binits{M.}},
  \bauthor{\bsnm{Taddy},~\bfnm{M.}\binits{M.}},
  \bauthor{\bsnm{Rao},~\bfnm{J.}\binits{J.}},
  \bauthor{\bsnm{Liu},~\bfnm{J.}\binits{J.}},
  \bauthor{\bsnm{Lymberopoulos},~\bfnm{D.}\binits{D.}} \AND
  \bauthor{\bsnm{McAuley},~\bfnm{J.}\binits{J.}}
(\byear{2017}).
\btitle{Modeling Consumer Preferences and Price Sensitivities from Large-Scale
  Grocery Shopping Transaction Logs}.
In \bbooktitle{International World Wide Web Conference}.
\end{binproceedings}
\endbibitem

\bibitem[\protect\citeauthoryear{Wang and Blei}{2011}]{Wang2011}
\begin{binproceedings}[author]
\bauthor{\bsnm{Wang},~\bfnm{C.}\binits{C.}} \AND
  \bauthor{\bsnm{Blei},~\bfnm{D.~M.}\binits{D.~M.}}
(\byear{2011}).
\btitle{Collaborative topic modeling for recommending scientific articles}.
In \bbooktitle{Proceedings of the 17th ACM SIGKDD International Conference on
  Knowledge Discovery and Data Mining}.
\end{binproceedings}
\endbibitem

\bibitem[\protect\citeauthoryear{Wolpin}{1984}]{wolpin1984estimable}
\begin{barticle}[author]
\bauthor{\bsnm{Wolpin},~\bfnm{Kenneth~I}\binits{K.~I.}}
(\byear{1984}).
\btitle{An estimable dynamic stochastic model of fertility and child
  mortality}.
\bjournal{Journal of Political economy}
\bvolume{92}
\bpages{852--874}.
\end{barticle}
\endbibitem

\end{thebibliography}

\end{document}